\documentclass{article}

\usepackage{acra}
\usepackage{amsmath,amssymb}
\usepackage{booktabs}
\usepackage{graphicx}
\usepackage{xcolor}
\usepackage{url}
\usepackage[percent]{overpic}
\usepackage{float}
\usepackage{placeins}
\usepackage{tikz}
\usetikzlibrary{arrows.meta,positioning,backgrounds}
\usepackage{pgfplots}
\pgfplotsset{compat=1.18}
\graphicspath{{./}}
\DeclareRobustCommand{\citep}[1]{\cite{#1}}
\newcommand{\keywords}[1]{\par\smallskip\noindent\textbf{Keywords:} #1\par\smallskip}

\title{HOC2R: Hand-Object Contact Trajectories for Human-to-Robot Motion Transfer}
\author{Hongkun Yu$^{1}$ \quad Francisco M. L\'opez$^{1}$ \quad
Francisco Cruz$^{1,2}$\\
{\normalsize $^{1}$School of Computer Science and Engineering, UNSW Sydney, Australia}\\
{\normalsize $^{2}$Escuela de Ingenier\'ia, Universidad Central de Chile, Santiago, Chile}\\
{\normalsize \texttt{z5669162@ad.unsw.edu.au, fm.lopez@unsw.edu.au, f.cruz@unsw.edu.au}}\\
{\normalsize Corresponding author: Hongkun Yu}}

\makeatletter
\def\@maketitle{\vbox to \titlebox{\hsize\textwidth
 \linewidth\hsize
 {\small\bfseries\raggedright
 Cite as: Hongkun Yu, Francisco M. L\'opez, and Francisco Cruz.
 HOC2R: Hand-Object Contact Trajectories for Human-to-Robot Motion Transfer.
 \textit{Proceedings of the Australasian Conference on Robotics and Automation
 (ACRA), 2026.}\par}
 \vskip 8pt
 \hrule height 0.4pt
 \vskip 10pt
 \centering
 {\LARGE\bf \@title \par} \vskip 0.2in plus 1fil minus 0.1in
 {\def\and{\unskip\enspace{\rm and}\enspace}%
  \def\And{\end{tabular}\hss \egroup \hskip 1in plus 2fil
           \hbox to 0pt\bgroup\hss \begin{tabular}[t]{c}\Large\bf}%
  \def\AND{\end{tabular}\hss\egroup \hfil\hfil\egroup
          \vskip 0.25in plus 1fil minus 0.125in
           \hbox to \linewidth\bgroup\Large \hfil\hfil
             \hbox to 0pt\bgroup\hss \begin{tabular}[t]{c}\Large\bf}%
  \hbox to \linewidth\bgroup\Large \hfil\hfil
    \hbox to 0pt\bgroup\hss \begin{tabular}[t]{c}\Large\bf\@author
                            \end{tabular}\hss\egroup
    \hfil\hfil\egroup}
  \vskip 0.3in plus 2fil minus 0.1in
}}
\makeatother

\begin{document}
\frenchspacing
\raggedbottom

\maketitle

\begin{abstract}
Human videos contain abundant object-manipulation demonstrations, but they do
not directly provide motion cues that a robot can follow. We present HOC2R, a
\textbf{H}and-\textbf{O}bject \textbf{C}ontact trajectory extractor for
\textbf{R}obot motion that combines frozen DINOv3 and SAM~2.1 features with
learned heads for hand-object association and contact localization. HOC2R
supports two modes: count-aware automation that detects, counts, and
localizes visible contacts in untrimmed videos, and fixed-two localization for
two-finger grippers when contact frames are known.
On a 30-task point-annotated benchmark, count-aware
automation improves contact-count accuracy from 58.1\% to 71.4\% and
reduces mean localization distance from 19.81 to 10.46 pixels relative to an
adapted VRB2 baseline. On seven unseen RGB-D rollouts, a five-task fixed-two
checkpoint reduces mean localization distance
from 67.67 to 27.02 pixels.
Across six robot evaluation sequences, the 30-task fixed-two checkpoint produces
four feasible trajectories and succeeds on three tasks, compared with two
feasible trajectories and one successful task for each adapted baseline. HOC2R is
competitive in the evaluated single-hand scenes, robust to the
hand-color shift, and strongest in the evaluated multi-hand scenes. These
results support HOC2R as a practical pseudo-labeling method for extracting
robot-usable contact trajectories from human demonstrations.
\end{abstract}


\section{Introduction}

Large-scale robot learning is limited by the cost of collecting diverse robot
demonstrations, while human videos contain abundant examples of grasping and
object manipulation. Prior work learns from human play, predicts visual affordances,
or transfers motion across embodiments
\citep{wang2023mimicplay,bahl2023affordances,xie2025human2robot}. A persistent
gap is the interaction geometry connecting observed human demonstrations to
robot actions. Boxes and masks identify participants, but not the contact
locations needed to derive a robot-gripper trajectory.

We study \emph{hand-object contact trajectories}: visible contacts linked
across frames of a human demonstration. They provide an interaction-centered
cue for downstream robot learning without requiring paired robot actions.
Extraction is difficult because fingertips are small or occluded, background
corners produce competing responses, and multiple hands and objects create
association ambiguity. Changes in hand color and appearance, viewpoint,
illumination, background, and camera setup further challenge transfer from
limited point annotations.

HOC2R addresses these failures by combining complementary frozen visual
representations. A proposal head over DINOv3 features identifies likely hand-object interaction
regions, while a learned heatmap head predicts contact likelihood from
whole-image SAM~2.1 features. Proposal-box touch scores rerank heatmap candidates. Count-aware
automation distinguishes contact from no-contact frames and predicts how many
contacts are visible. For downstream robot control with a two-finger gripper,
fixed-two localization assumes that consecutive frames containing hand-object
contact have already been identified. We call these frames the active contact
segment. Fixed-two then returns two proposal-constrained points. This separation is
important: count-aware automation is strongest on held-out in-domain tasks,
while fixed-two localization is more
reliable after domain shift.

HOC2R provides a practical starting checkpoint for extracting robot-usable
contact trajectories from human manipulation videos. Its principal advantage
appears under multi-hand ambiguity, where adapted baselines can follow a
resting or non-manipulating hand. We further test whether the extracted
contacts remain useful for robot control after RGB-D lifting to 3D and
geometric safety filtering.

Our contributions are:
\begin{itemize}
    \item HOC2R, a hand-object contact-trajectory extractor that combines
    proposal geometry learned from DINOv3 features with point localization
    learned from SAM~2.1 features for count-aware automation and fixed-two
    localization.
    \item evaluations across held-out tasks, domains, and model components that expose
    multi-hand robustness, domain sensitivity of contact-presence and contact-count prediction, and the effect of
    small but diverse adaptation sets.
    \item calibrated RGB-D robot evaluation showing that the extracted
    contacts can produce feasible post-contact trajectories for grasping,
    pushing, and two-hand pen motion.
\end{itemize}
Code and trained checkpoints are publicly available.\footnote{Project
repository: \url{https://github.com/liszzzzt/HOC2R}.}

\section{Related Work}

Learning from human demonstrations offers a path toward scaling robot learning
beyond trajectories collected directly on each target platform. Prior work has
studied which properties of offline demonstrations most affect manipulation
learning, showing that demonstration quality, observation design, and training
choices can matter as much as dataset size \citep{mandlekar2021what}.
MimicPlay uses human play to guide long-horizon imitation
\citep{wang2023mimicplay}, while ImitationNet retargets human motion across
embodiments \citep{yan2023imitationnet}. Paired human-robot data further support
direct action prediction in Human2Robot \citep{xie2025human2robot}, and ATM
learns arbitrary point trajectories from video \citep{wen2024atm}. These methods
address policy learning, motion retargeting, or cross-embodiment action
prediction. HOC2R instead extracts visible contact locations from human
demonstrations without paired robot actions.

Affordance and hand-object understanding provide complementary descriptions
of interaction. AffordanceNet predicts pixel-level functional regions of
objects, associating image areas with semantic actions such as grasping or
cutting \citep{do2018affordancenet}. At the instance level, 100-DOH detects hands,
their contact states, and associated objects in unconstrained images
\citep{shan2020hands}. EgoHOS provides hand, interacting-object, and
contact-boundary segmentation \citep{zhang2022egohos}. Together, these works show
that contact can be represented through affordance categories, detections,
states, masks, and boundaries. These outputs are not interchangeable
with contact-point trajectories observed in a particular demonstration.

VRB is the closest baseline to our objective because it learns contact
affordances and post-contact motion cues from human videos
\citep{bahl2023affordances}. Its formulation is designed to predict actionable
future behavior, whereas HOC2R extracts visible contacts from each observed
frame. HOWTransfer reconstructs hand motion and contact intervals for
grasp retargeting, while ContactFlow uses 3D contact trajectories to condition a
cross-embodiment world model
\citep{shi2026howtransfer,azirar2026contactflow}. HOC2R instead learns visible
2D contact points from limited point supervision and lifts their midpoint
trajectory into robot-frame 3D for post-contact motion.

We adapt VRB-style hand-object intersections and EgoHOS contact boundaries to
point outputs so that all methods can be evaluated under the same localization
and robot-trajectory protocols. HOC2R instead trains lightweight point-supervised
heads over frozen DINOv3 and SAM~2.1 features
\citep{simeoni2025dinov3,ravi2024sam2}. This design follows evidence that
reusable human-video representations and diverse visual pretraining can support
data-efficient embodied learning
\citep{nair2023r3m,majumdar2023vc1}. Our
experiments ask whether such representations can produce contact trajectories
that remain useful after domain shift, RGB-D lifting, and robot execution.

\section{Method}

\subsection{Contact Representation and Shared Architecture}

Given an RGB frame $I_t$, HOC2R predicts an unordered set of visible
hand-object contacts
\begin{equation}
    \widehat{\mathcal C}_t
    =\{(\hat x_{t,i},\hat y_{t,i})\}_{i=1}^{\hat n_t},
    \qquad \hat n_t\in\{0,1,2\}.
\end{equation}
We target motion transfer to a two-finger gripper and select demonstrations
performed with two contact fingers. The contact count $\hat n_t$ is the number
of visible contact points. Zero denotes no contact, one denotes that one of the
two contacts is occluded, and two denotes that both contacts are visible. Across frames, the
predicted sets define an
image-space contact trajectory.

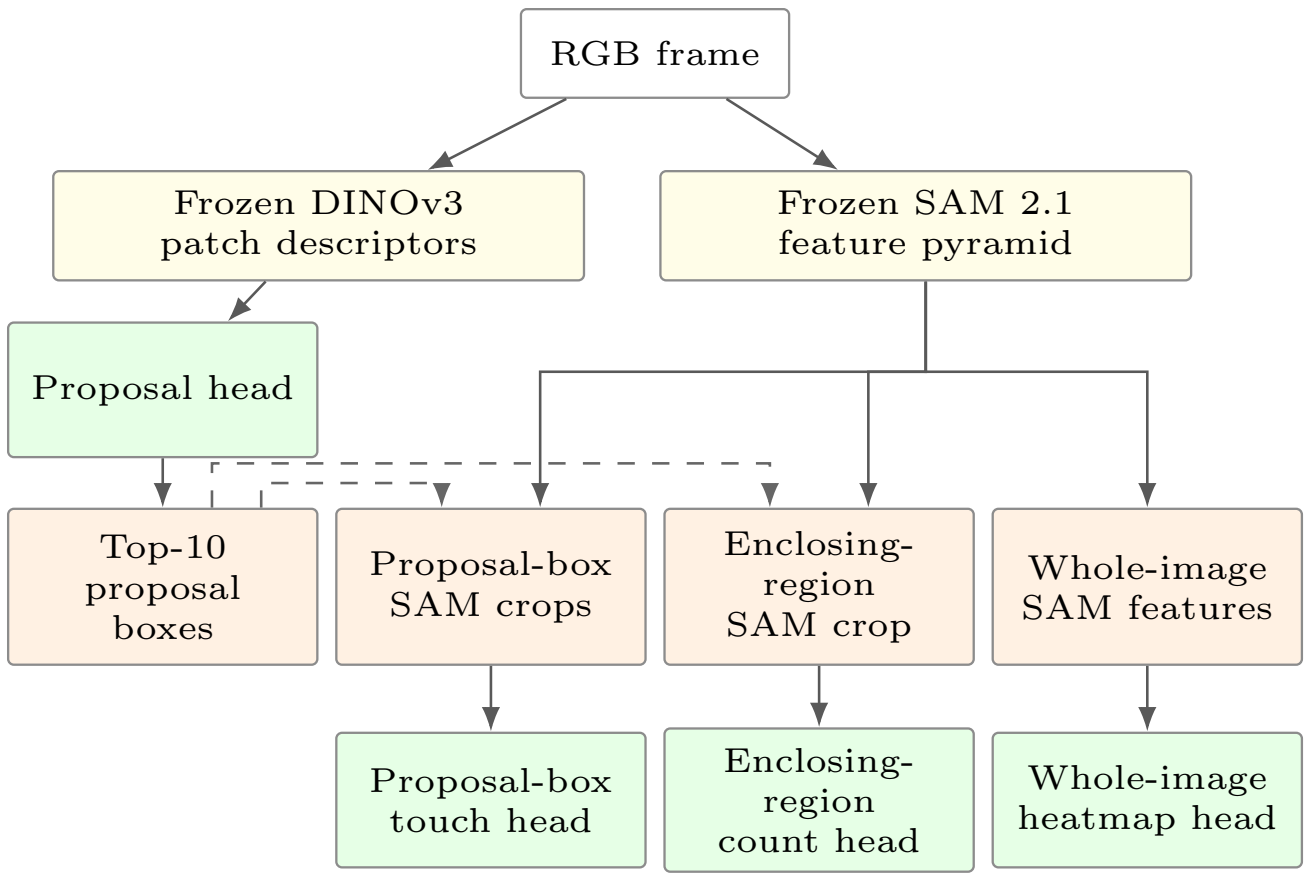
\begin{figure}[!t]
\centering
\resizebox{\columnwidth}{!}{%
\begin{tikzpicture}[
  >={Latex[length=1.5mm]},
  flow/.style={->,draw=black!65,line width=0.45pt},
  geometry/.style={->,draw=black!60,dashed,line width=0.45pt},
  cell/.style={draw=black!45,rounded corners=1pt,align=center,
               minimum height=0.54cm,font=\tiny},
  frozen/.style={cell,fill=yellow!11,text width=3.00cm,
                 minimum height=0.62cm},
  operation/.style={cell,fill=orange!11,text width=1.65cm,
                    minimum height=0.95cm},
  learned/.style={cell,fill=green!10,text width=1.65cm,
                  minimum height=0.82cm}
]
  \node[cell,text width=1.40cm] (rgb) at (0,0) {RGB frame};

  \node[frozen] (dino) at (-2.05,-1.05) {Frozen DINOv3 patch descriptors};
  \node[frozen] (sam) at (1.65,-1.05) {Frozen SAM~2.1 feature pyramid};

  \node[learned] (proposal) at (-3.0,-2.05) {Proposal head};

  \node[operation] (boxes) at (-3.0,-3.25)
    {Top-10\\proposal boxes};
  \node[operation] (boxcrop) at (-1.0,-3.25)
    {Proposal-box\\SAM crops};
  \node[operation] (enclose) at (1.0,-3.25)
    {Enclosing-region\\SAM crop};
  \node[operation] (whole) at (3.0,-3.25)
    {Whole-image\\SAM features};

  \node[learned] (touch) at (-1.0,-4.55)
    {Proposal-box\\touch head};
  \node[learned] (count) at (1.0,-4.55)
    {Enclosing-region\\count head};
  \node[learned] (heat) at (3.0,-4.55)
    {Whole-image\\heatmap head};

  \draw[flow] (rgb) -- (dino);
  \draw[flow] (rgb) -- (sam);
  \draw[flow] (dino) -- (proposal);
  \draw[flow] (proposal) -- (boxes);

  \begin{scope}[on background layer]
    \draw[flow] (sam.south) -- ++(0,-0.55) -| ([xshift=3mm]boxcrop.north);
    \draw[flow] (sam.south) -- ++(0,-0.55) -| ([xshift=3mm]enclose.north);
    \draw[flow] (sam.south) -- ++(0,-0.55) -| (whole.north);
    \draw[geometry] ([xshift=6mm]boxes.north) -- ++(0,0.15)
      -| ([xshift=-3mm]boxcrop.north);
    \draw[geometry] ([xshift=3mm]boxes.north) -- ++(0,0.27)
      -| ([xshift=-3mm]enclose.north);
  \end{scope}

  \draw[flow] (boxcrop) -- (touch);
  \draw[flow] (enclose) -- (count);
  \draw[flow] (whole) -- (heat);
\end{tikzpicture}%
}
\caption{HOC2R training architecture. The proposal head uses frozen DINOv3
features and patch labels to predict candidate contact-region boxes. The
proposal-box touch head uses each proposal's SAM feature crop and box label to
predict whether the box contains contact. The enclosing-region count head uses
the SAM feature crop covering the top-ten proposal boxes and frame-level labels
to predict contact presence and count. The whole-image heatmap head uses SAM
features and Gaussian targets to predict pixel-wise contact likelihood.}
\label{fig:method_training_v1}
\vspace{-6pt}
\end{figure}

HOC2R combines frozen DINOv3 ViT-S/16 features
\citep{simeoni2025dinov3} and SAM~2.1 Hiera-S features
\citep{ravi2024sam2}. Four lightweight heads are trained from features.
The proposal head uses DINOv3 features to return the ten highest-scoring
$32\times32$ contact-region proposal boxes.
Their smallest axis-aligned bounding rectangle defines the enclosing region
for the count head.
Three heads use SAM features: proposal-box touch predicts whether each box
contains contact, enclosing-region count predicts contact presence followed by
one-versus-two classification, and a learned whole-image heatmap predicts
pixel-wise contact likelihood. Figure~\ref{fig:method_training_v1} summarizes
the training architecture.

\subsection{Count-Aware Automation and Fixed-Two Localization}

HOC2R supports two deployment modes sharing proposal boxes, heatmap candidates,
and proposal-box touch predictions. Count-aware automation handles untrimmed
sequences by predicting contact presence and count. Fixed-two
localization assumes a known active contact segment and always returns two
points for a two-finger gripper.

Both modes extract local maxima from the contact-likelihood heatmap. Let $h(p)$ be the
probability at candidate $p$, and let $q(p)$ be the largest touch
probability among proposal boxes containing $p$, with $q(p)=0.5$ outside all
proposal boxes. The shared proposal-box touch-score soft reranking is
\begin{equation}
    s(p)=\operatorname{logit}h(p)
    +0.5\max\!\left(0,\operatorname{logit}q(p)\right).
\end{equation}
The globally fixed weight is selected on validation folds. Because the touch
term is clipped at zero, proposal-box touch scores above 0.5 can raise a
candidate's rank, whereas lower scores do not penalize it.

\begin{figure}[H]
\centering
\input{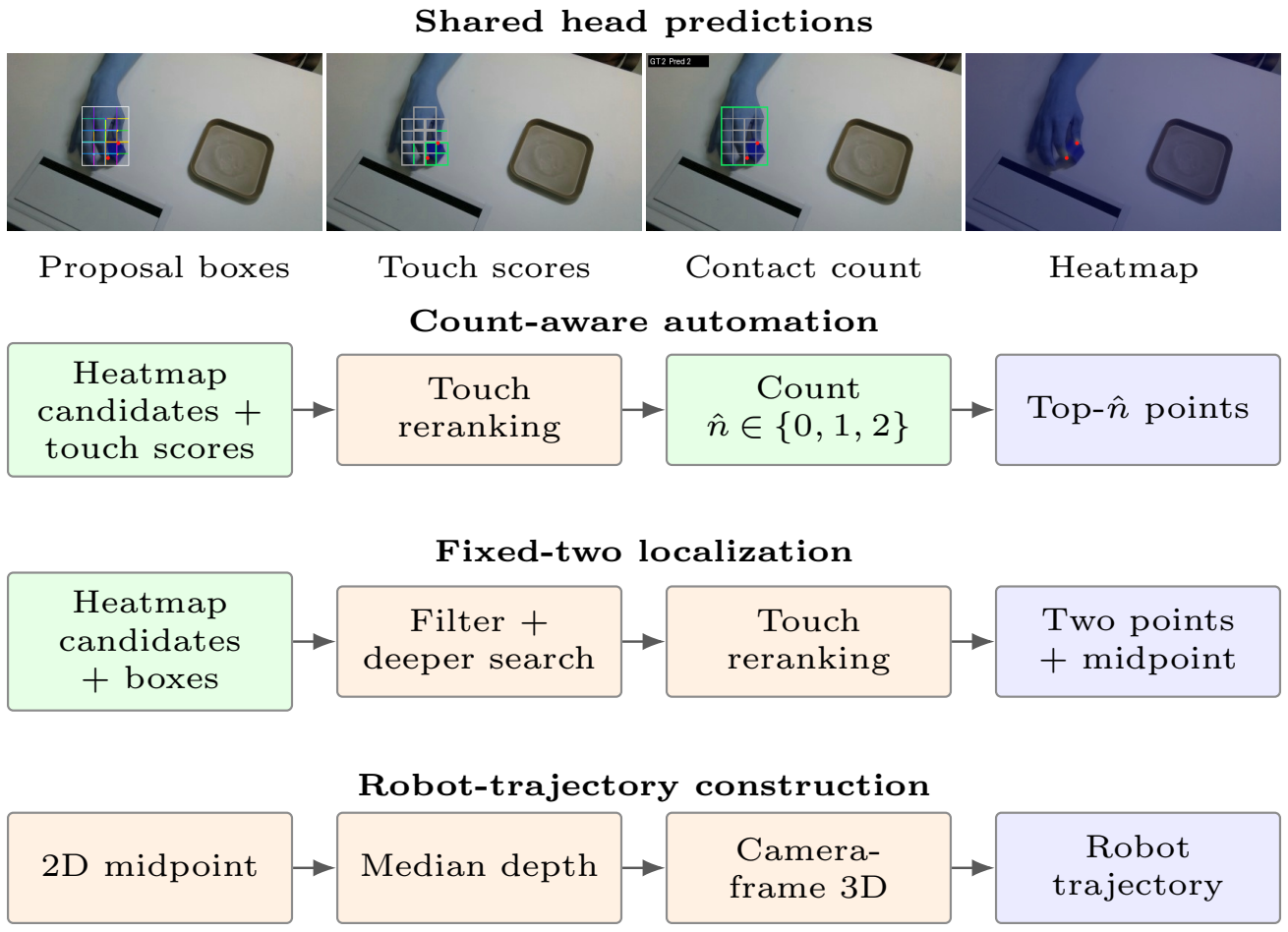}
\caption{HOC2R inference architecture. Shared outputs comprise
contact-region proposal boxes, per-box touch scores, an enclosing-region
contact-count prediction, and a whole-image contact-likelihood heatmap.
Count-aware automation reranks heatmap candidates using touch scores and emits
the predicted number of points. Fixed-two localization filters candidates
through proposal boxes, progressively examines more heatmap maxima until two
eligible candidates are found, applies touch reranking, and returns two points
for a known active contact segment. Their midpoint is lifted with aligned depth
and camera calibration into a robot-base trajectory.}
\label{fig:method_inference_v1}
\end{figure}

\noindent\textbf{Count-aware automation} is intended for untrimmed video. The
enclosing-region head first predicts contact presence and then the contact
count, one or two, for positive frames. The decoder emits the corresponding number of
reranked heatmap peaks.

\noindent\textbf{Fixed-two localization} is intended for a
known active contact segment. It bypasses counting, requires candidates to lie
inside individual proposal boxes, and returns the two highest-ranked
points. The fixed-two mode therefore prioritizes an uninterrupted sequence of
spatially constrained predictions over automatic onset and release detection.
Figure~\ref{fig:method_inference_v1} summarizes the two decoding modes and
their shared robot-trajectory construction.

\subsection{From Image Contacts to Robot Motion}

For transferring predicted contacts to robot motion, we compute the 2D
midpoint before depth lookup,
sample the median valid aligned depth in a $3\times3$ neighborhood, deproject
the point with RealSense intrinsics, and transform it into the Kinova
manipulator's base frame using calibrated eye-to-hand geometry. A shared offline filter flags
missing depth, workspace violations, and isolated geometric jumps. Short gaps
are interpolated, accepted trajectories are resampled to bounded waypoint spacing,
and trajectories exceeding fixed repair limits are rejected before inverse-kinematics
execution. Every method uses the same calibration, filter, workspace, and
controller. Calibration, workspace limits, and trajectory-repair criteria are
detailed in Appendix~\ref{app:rgbd_robot_acra}.

\section{Experimental Setup}

\paragraph{Annotated benchmark.}
We annotate 210 blue-hand human-view frames from 30 Human2Robot tasks
\citep{xie2025human2robot}. These comprise 87 no-contact frames, 39
with one visible contact, and 84 with two visible contacts, totaling 207 visible points.
The tasks span grasping, pushing, pulling, writing, cloth handling, and
related task variants, while the selected frames cover start, pre-contact,
first-contact, motion, placement, and release stages. Evaluation uses five
task-level folds, with complete tasks assigned to training, validation, or
testing. In each round, 18 training tasks fit the lightweight heads, six
validation tasks select checkpoints and thresholds, and six held-out tasks
provide the final evaluation.
This prevents visually
similar frames from one demonstration from crossing the train-test boundary.
The selected frames, point annotations, derived labels, and annotation tool
are publicly available.\footnote{Dataset repository:
\url{https://github.com/liszzzzt/HOC2R-Dataset}.}

\begin{figure}[H]
\centering
\begin{minipage}[c]{0.247\linewidth}\centering
  \includegraphics[width=\linewidth,height=2.15cm,keepaspectratio]{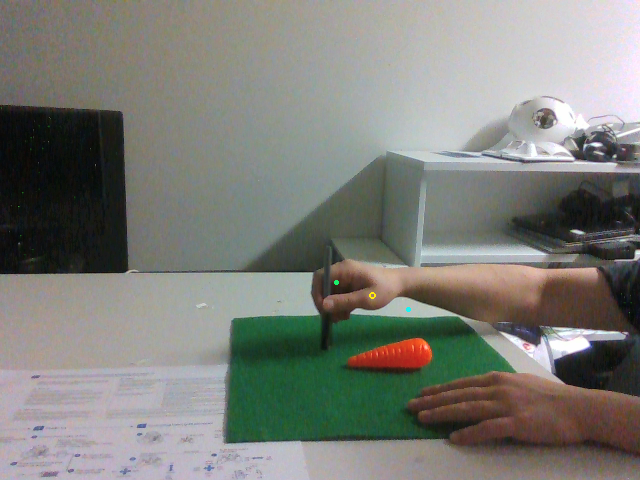}\\[-2pt]
  {\tiny (a) Contacts}
\end{minipage}\hfill
\begin{minipage}[c]{0.247\linewidth}\centering
  \includegraphics[width=\linewidth,height=2.15cm,keepaspectratio]{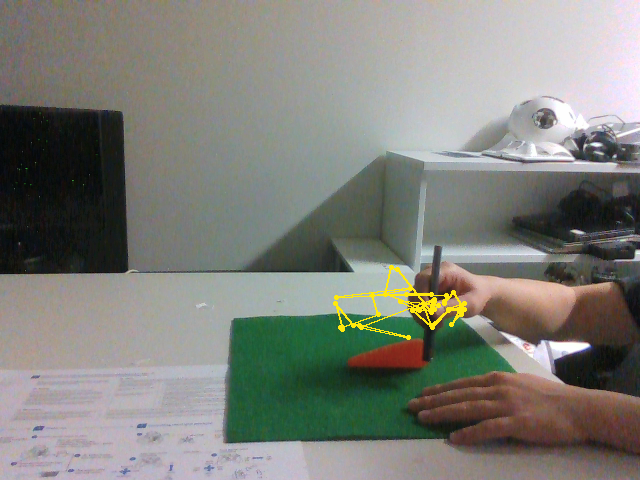}\\[-2pt]
  {\tiny (b) Trajectory}
\end{minipage}\hfill
\begin{minipage}[c]{0.247\linewidth}\centering
  \includegraphics[width=\linewidth,height=2.15cm,keepaspectratio]{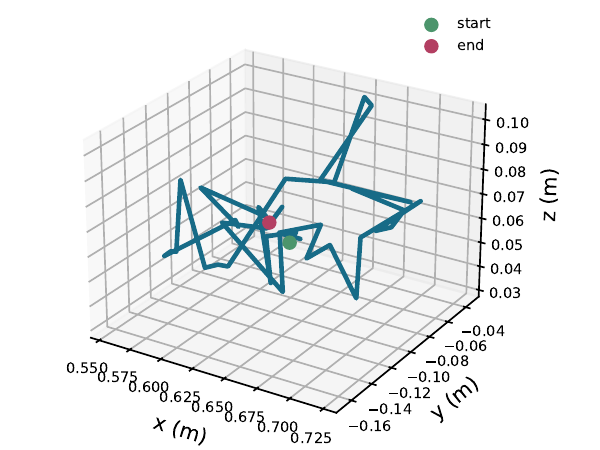}\\[-2pt]
  {\tiny (c) Lift to 3D}
\end{minipage}\hfill
\begin{minipage}[c]{0.247\linewidth}\centering
  \includegraphics[width=\linewidth,height=2.15cm,keepaspectratio]{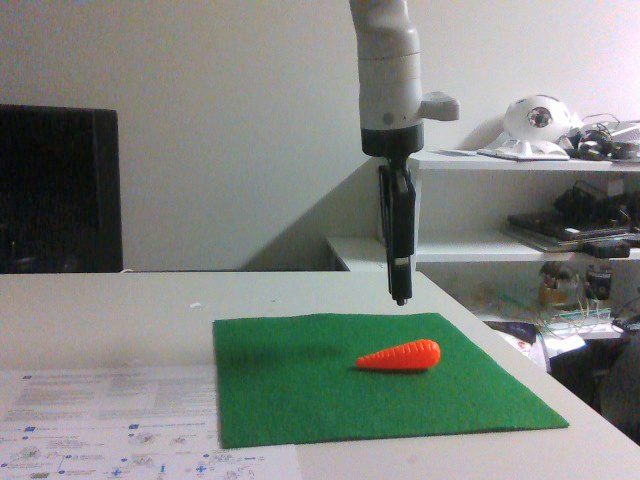}\\[-2pt]
  {\tiny (d) IK}
\end{minipage}
\caption{Robot evaluation workflow illustrated on the pen-motion
demonstration. HOC2R extracts two image contacts (a), whose midpoint forms a
2D trajectory (b). Aligned depth and camera calibration map the trajectory
into the robot base frame (c). Shared filtering and waypoint resampling are
applied before inverse-kinematics execution (d).}
\label{fig:robot_workflow_v1}
\end{figure}

\paragraph{External and robot evaluation.}
Unlike the annotated blue-hand benchmark, external evaluation uses 207
annotated frames from seven RGB-D rollouts with natural-skin hands, unseen
objects, and changes in viewpoint, background, and lighting.
The pencil rollout contains two visible hands. No external frame is used for
training. Robot evaluation uses six evaluation sequences in a cluttered
lab: two grasp-and-place sequences, two pushing sequences, and two pen-motion
sequences with two visible hands.
Each sequence starts from established gripper contact, so it evaluates
post-contact motion rather than autonomous grasp acquisition.
Figure~\ref{fig:robot_workflow_v1} summarizes the evaluation workflow.

\paragraph{Baselines.}
We compare HOC2R with two adapted hand-object contact baselines. We refer to
our two-point adaptation of the contact-extraction stage used by
VRB~\citep{bahl2023affordances} as VRB2. It
combines 100-DOH hand and object detections~\citep{shan2020hands} with
hand-color segmentation, then reduces the resulting hand-object boundary
intersection to two centers using a weighted Gaussian mixture model.
Similarly, EgoHOS-Point is our two-point adaptation of the pretrained EgoHOS
hand-contact boundary predictor~\citep{zhang2022egohos}. It converts the
predicted boundary into two representative points. Neither baseline is
retrained on our point labels. We evaluate their released pretrained
predictions through fixed point-output adapters and apply the same localization,
trajectory-filtering, and robot-control protocols to all methods. This setup
measures practical transfer to unseen external and robot data.

\paragraph{Evaluation metrics.}
We measure contact presence, contact count, and localization.
Contact-presence accuracy measures whether a frame contains contact, while
contact-count accuracy requires the predicted count $0/1/2$ to match the
annotation. Point $F1@10$ uses one-to-one matching and counts a predicted
point as correct when it lies within 10 pixels of an annotation. Conditional
mean and median localization distance are computed only when a positive frame
has both an annotation and an emitted prediction, so we also report missed
positive frames and false contact on no-contact frames. Confidence intervals
are estimated by resampling complete tasks 10,000 times with replacement and
recomputing each metric. For robot evaluation, a trajectory is feasible if
it passes the shared pre-execution filter and can therefore be executed safely.
Task success requires the executed motion to achieve the predefined task
objective. Rejected trajectories are counted as task failures.

\section{Results}

We organize the results around four questions: in-domain contact prediction,
sensitivity to training-task count, zero-shot external transfer, and utility
for robot motion.

\subsection{RQ1: How Does HOC2R Compare with Contact-Point Baselines?}

On the held-out in-domain benchmark, the complete count-aware model is
the strongest overall HOC2R configuration. Table~\ref{tab:indomain_v1} reports
results across the five held-out folds. HOC2R count-aware automation improves contact-count accuracy from
58.1\% to 71.4\% over VRB2 and raises
$F1@10$ from 25.8\% to 62.0\%. Its mean localization distance is 10.46 pixels
versus 19.81 pixels for VRB2. EgoHOS-Point has no in-domain result because its
pretrained hand detector returns no prediction for any blue-hand frame.
Fixed-two predicts neither contact presence nor contact count, so both
accuracies are undefined. Its point scores also include predictions on no-contact frames
because it always emits two points.

\begin{table}[H]
\centering
\footnotesize
\setlength{\tabcolsep}{2.5pt}
\caption{Held-out in-domain results for contact presence, contact count, and
localization. Dashes denote no EgoHOS-Point output or undefined fixed-two
contact metrics.}
\label{tab:indomain_v1}
\resizebox{\columnwidth}{!}{%
\begin{tabular}{@{}lrrrr@{}}
\toprule
Method & Pres. $\uparrow$ & Count $\uparrow$ & $F1@10$ $\uparrow$ &
Mean/med. $\downarrow$ \\
\midrule
VRB2 & 75.2 & 58.1 & 25.8 & 19.81 / 15.08 \\
EgoHOS-Point & -- & -- & -- & -- \\
HOC2R count-aware & \textbf{83.3} & \textbf{71.4} & \textbf{62.0} &
\textbf{10.46 / 4.81} \\
HOC2R fixed-two & -- & -- & 51.4 & 11.37 / 5.66 \\
\bottomrule
\end{tabular}%
}
\end{table}

Table~\ref{tab:ablations_v1} isolates the decoding components while keeping
the whole-image heatmap fixed. Hard proposal filtering gives the lowest
conditional mean distance but can discard valid candidates. Touch reranking
improves $F1@10$ over the heatmap-only decoder, while the complete count-aware
automation achieves the best $F1@10$ and median distance. Fixed-two is lower on this
untrimmed benchmark because it also emits points on no-contact frames.

\begin{table}[H]
\centering
\footnotesize
\setlength{\tabcolsep}{3pt}
\caption{Component and decoder ablations on the held-out in-domain frames.}
\label{tab:ablations_v1}
\begin{tabular}{@{}lrrr@{}}
\toprule
Variant & $F1@10$ $\uparrow$ & Mean $\downarrow$ & Median $\downarrow$ \\
\midrule
Heatmap only (0/2) & 59.5 & 12.05 & 5.03 \\
Hard proposal filter (0/2) & 59.3 & \textbf{10.14} & 4.87 \\
Touch reranking (0/2) & 61.8 & 10.73 & 4.91 \\
HOC2R count-aware & \textbf{62.0} & 10.46 & \textbf{4.81} \\
HOC2R fixed-two & 51.4 & 11.37 & 5.66 \\
\bottomrule
\end{tabular}
\end{table}

Before combining the heads, we examine proposal coverage as retained
proposal boxes increase. We select the top ten as a practical
balance: together, these boxes cover 95.7\% of held-out contact annotations
while keeping the candidate set compact. This strong coverage provides a
useful interaction prior without cropping the whole-image SAM features. A
separate advantage appears in contact-count prediction. HOC2R explicitly
predicts all three contact counts (0, 1, and 2), whereas the adapted VRB2 point decoder
is restricted to zero or two outputs. Contact-count confusion is reported in
Appendix~\ref{app:contact_count_acra}.

\subsection{RQ2: How Does Training-Task Count Affect the Two Modes?}

The controlled data-efficiency study keeps five validation and five test tasks
fixed while expanding a cumulative training set from 5 to 20 tasks, so each
larger set contains the preceding one. The validation and test tasks were
chosen to span diverse objects and manipulation types. In-domain performance
generally improves with additional tasks. Count-aware $F1@10$ rises from
40.0\% to 56.0\%, while fixed-two $F1@10$ rises from 44.2\% to 50.0\%. More
training tasks therefore benefit both modes on the target distribution.
Table~\ref{tab:efficiency_v1} reports the results for the four
training subsets.

\begin{table}[!htbp]
\centering
\footnotesize
\setlength{\tabcolsep}{3pt}
\caption{Data-efficiency results on fixed in-domain and external evaluation
sets.}
\label{tab:efficiency_v1}
\begin{tabular}{@{}rlrr@{}}
\toprule
Train tasks & Evaluation mode & $F1@10$ $\uparrow$ & Mean dist. $\downarrow$ \\
\midrule
5 & In-domain count-aware & 40.0 & 16.05 \\
5 & In-domain fixed-two & 44.2 & 13.27 \\
5 & External fixed-two & \textbf{32.8} & \textbf{27.02} \\
\addlinespace[1pt]
10 & In-domain count-aware & 40.0 & 18.44 \\
10 & In-domain fixed-two & 46.2 & 11.78 \\
10 & External fixed-two & 26.0 & 30.46 \\
\addlinespace[1pt]
15 & In-domain count-aware & 51.1 & 15.27 \\
15 & In-domain fixed-two & 48.1 & 14.30 \\
15 & External fixed-two & 24.8 & 31.47 \\
\addlinespace[1pt]
20 & In-domain count-aware & \textbf{56.0} & \textbf{12.03} \\
20 & In-domain fixed-two & \textbf{50.0} & \textbf{11.75} \\
20 & External fixed-two & 24.8 & 28.27 \\
\bottomrule
\end{tabular}
\end{table}

Performance on the external rollouts follows a different pattern. Among the
four training subsets, the five-task fixed-two checkpoint performs best,
reaching 32.8\% $F1@10$ with a 27.02-pixel mean localization distance. The
20-task checkpoint performs worse, reaching 24.8\% $F1@10$ with a 28.27-pixel
mean localization distance. Because the subsets differ in both size and task composition and
only one training seed is available, these results do not establish that less
data is generally better. They instead indicate that task diversity and domain
composition are important when adapting frozen visual representations.

\subsection{RQ3: Does HOC2R Transfer to New Scenes and Multi-Hand Interactions?}

External evaluation includes several domain shifts, with multi-hand scenes
providing the most important change absent from training. Although none of the
annotated training frames contains a multi-hand scene, HOC2R remains effective
under this shift and can distinguish the manipulating interaction from resting
or non-manipulating hands.

\begin{figure}[H]
\centering
\includegraphics[width=\linewidth]{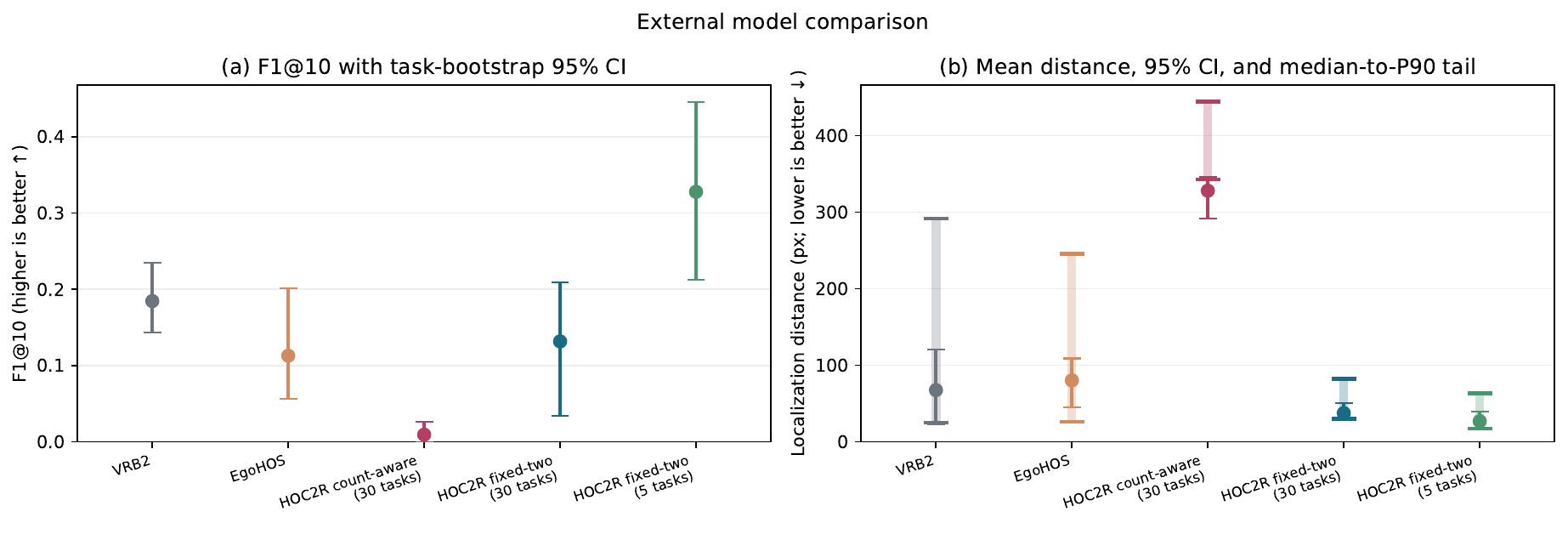}
\caption{Comparison on seven external RGB-D rollouts. In (a), points show
aggregate $F1@10$, and thin bars show task-level bootstrap 95\% confidence
intervals. In (b), points show mean localization distance, thin bars show its
95\% confidence interval, and wide stems span the median-to-90th-percentile
error. Shorter stems indicate more consistent localization.}
\label{fig:external_metric_uncertainty_v1}
\end{figure}

Diagnostic outputs show that the proposal boxes still identify
plausible interaction regions and that the learned whole-image heatmap retains
useful contact-point candidates. However, contact-presence prediction often
suppresses these candidates, causing the 30-task count-aware checkpoint to miss
63 of 129 positive frames and reach only 0.9\% $F1@10$. This failure motivates
evaluating fixed-two localization when the active contact segment is known.
Fixed-two bypasses the unreliable cross-domain contact-presence and
contact-count decisions and evaluates whether the spatial components can still
localize contact.
We report aggregate performance in Figure~\ref{fig:external_metric_uncertainty_v1}.

\begin{figure}[!htb]
\centering

{\raggedright\textbf{VRB2}\par}\vspace{2pt}
\begin{minipage}[t]{.247\linewidth}\centering
  \includegraphics[width=\linewidth]{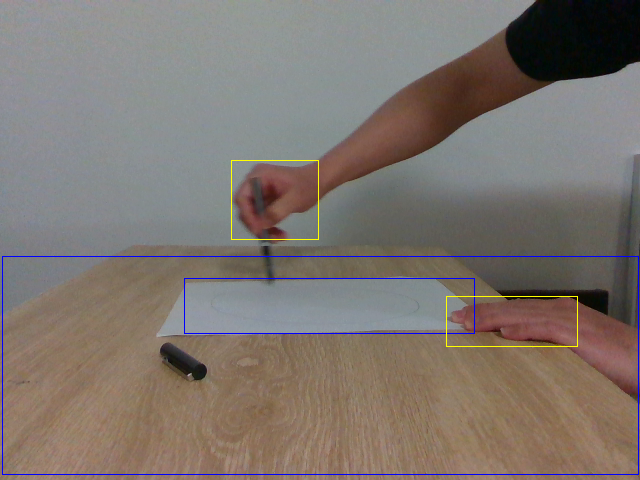}\\[-2pt]
  {\scriptsize 100-DOH}
\end{minipage}\hfill
\begin{minipage}[t]{.247\linewidth}\centering
  \includegraphics[width=\linewidth]{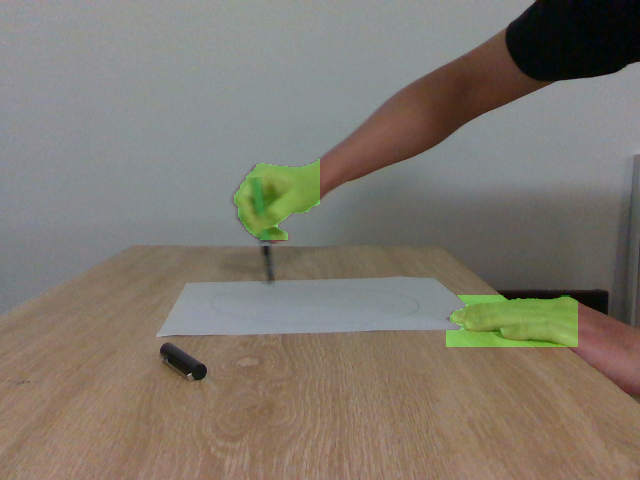}\\[-2pt]
  {\scriptsize Hand mask}
\end{minipage}\hfill
\begin{minipage}[t]{.247\linewidth}\centering
  \includegraphics[width=\linewidth]{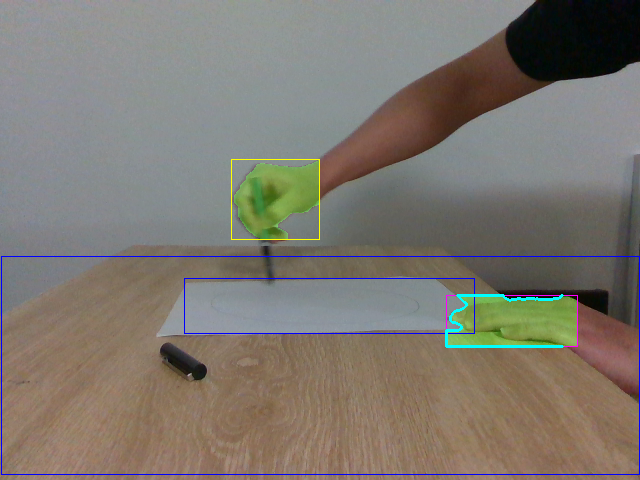}\\[-2pt]
  {\scriptsize Intersection}
\end{minipage}\hfill
\begin{minipage}[t]{.247\linewidth}\centering
  \includegraphics[width=\linewidth]{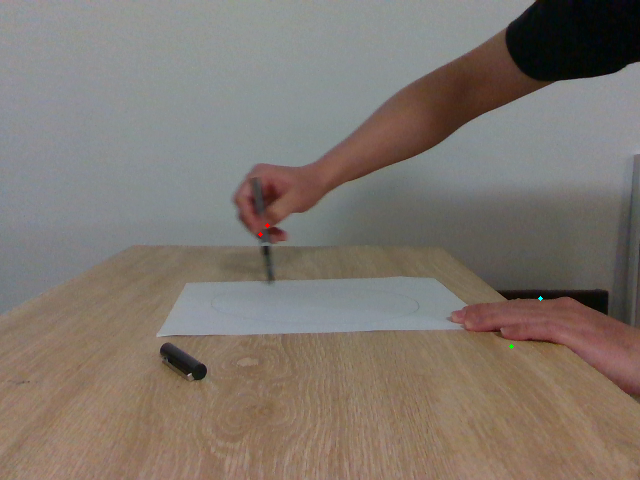}\\[-2pt]
  {\scriptsize Final points}
\end{minipage}

\vspace{2pt}
{\raggedright\textbf{EgoHOS-Point}\par}\vspace{2pt}
\makebox[\linewidth][l]{%
\begin{minipage}[t]{.247\linewidth}\centering
  \includegraphics[width=\linewidth]{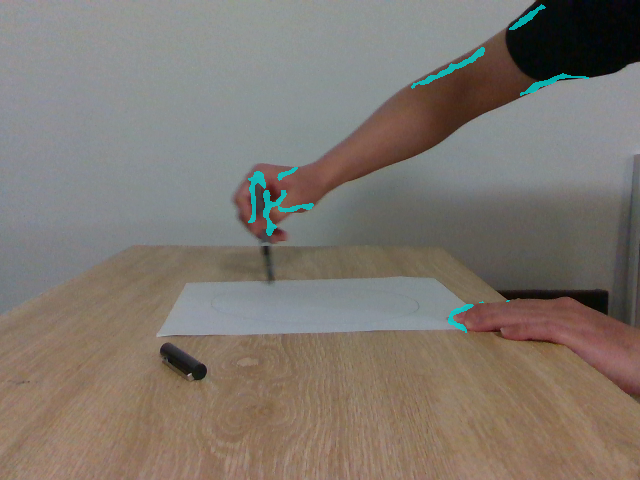}\\[-2pt]
  {\scriptsize Contact boundary}
\end{minipage}\hspace{.004\linewidth}
\begin{minipage}[t]{.247\linewidth}\centering
  \includegraphics[width=\linewidth]{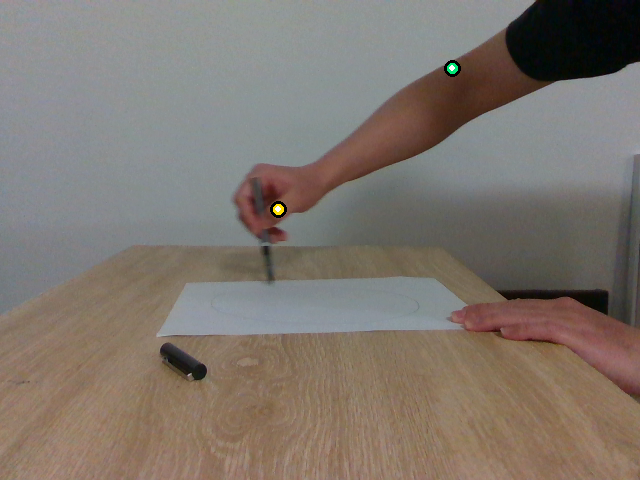}\\[-2pt]
  {\scriptsize Converted points}
\end{minipage}%
}

\vspace{2pt}
{\raggedright\textbf{HOC2R fixed-two (30 tasks)}\par}\vspace{2pt}
\begin{minipage}[t]{.247\linewidth}\centering
  \includegraphics[width=\linewidth]{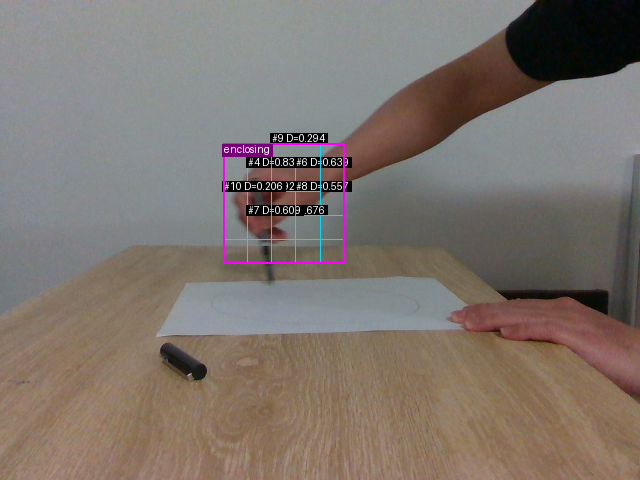}\\[-2pt]
  {\scriptsize Proposal boxes}
\end{minipage}\hfill
\begin{minipage}[t]{.247\linewidth}\centering
  \includegraphics[width=\linewidth]{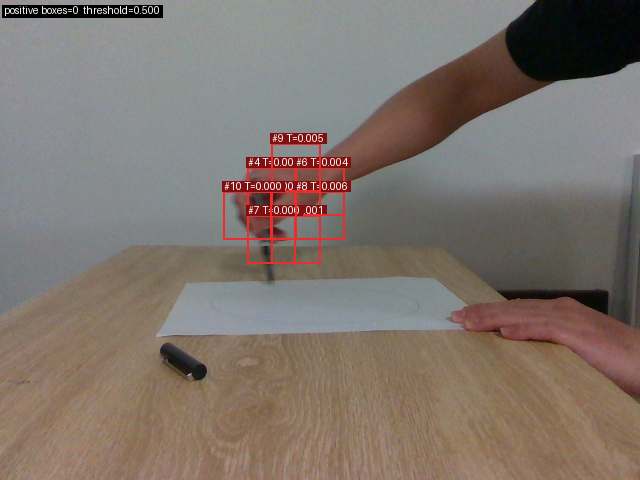}\\[-2pt]
  {\scriptsize Box touch}
\end{minipage}\hfill
\begin{minipage}[t]{.247\linewidth}\centering
  \includegraphics[width=\linewidth]{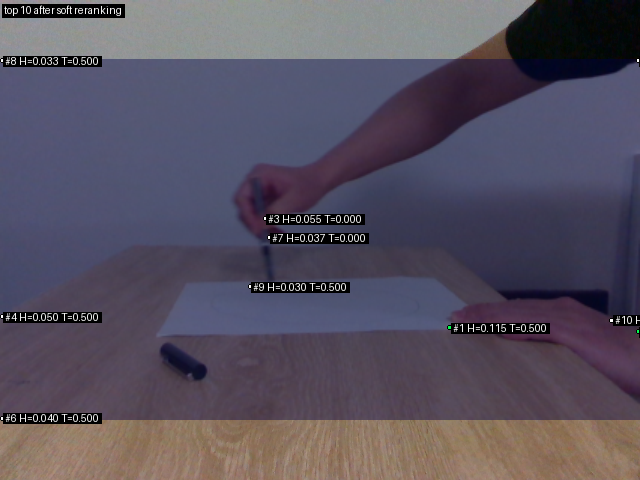}\\[-2pt]
  {\scriptsize Heatmap candidates}
\end{minipage}\hfill
\begin{minipage}[t]{.247\linewidth}\centering
  \includegraphics[width=\linewidth]{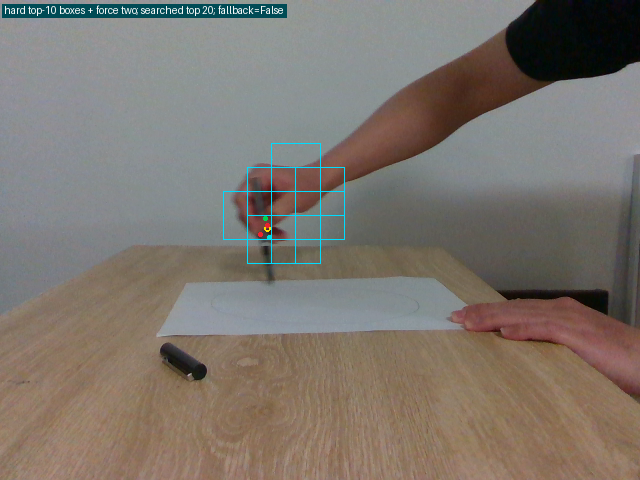}\\[-2pt]
  {\scriptsize Fixed-two points}
\end{minipage}

\caption{External multi-hand diagnostics. VRB2 associates its prediction with
the resting hand. EgoHOS responds to the resting hand and the clothing sleeve
beside the manipulating arm, and its point adapter retains these unrelated
boundaries. In contrast, HOC2R proposal boxes focus on the manipulating hand,
and the fixed-two decoder localizes the pencil contact. The assignment errors
are 156.11 px for VRB2, 126.00 px for EgoHOS-Point, and 17.81 px for HOC2R.}
\label{fig:external_pencil_diagnostic}
\end{figure}

RQ2 identifies the five-task checkpoint in fixed-two mode as the best-performing
checkpoint in the external data-efficiency study. We also evaluate the final 30-task
checkpoint trained on the full annotated dataset. The five-task checkpoint
achieves the highest aggregate $F1@10$ and lowest mean localization distance, outperforming
both adapted baselines. The 30-task checkpoint also gives lower mean localization distance
than both baselines, although its $F1@10$ is lower than VRB2. Fixed-two avoids
missed positive frames by construction, but assumes that contact frames are known
and therefore does not perform contact-presence or contact-count prediction.

Multi-hand robustness is also evident in the 30-task checkpoint. In the
two-hand pencil rollout, VRB2 and
EgoHOS-Point can associate contact with the resting hand. Despite having no
multi-hand training frames, HOC2R preserves
candidate regions around multiple hand-object interactions and assigns
stronger proposal scores to the manipulating contact. The 30-task proposal
boxes cover 68 of 77 annotated pencil contacts.
Figure~\ref{fig:external_pencil_diagnostic} shows this behavior.
On simpler single-hand tasks, HOC2R is competitive but not uniformly superior.
HOC2R leads on four of the seven rollouts. VRB2 gives lower mean localization distance on two
rollouts and is marginally better on tissue.
Complete per-rollout localization results are reported in
Appendix~\ref{app:external_per_rollout_acra},
Table~\ref{tab:external_tasks_acra}.

\subsection{RQ4: Do the Trajectories Support Robot Motion?}

The robot study asks whether the image-space advantage survives depth lifting,
camera-to-robot transformation, and geometric safety checks. Across six
robot evaluation sequences per method, the 30-task HOC2R fixed-two checkpoint produces
four feasible trajectories and succeeds on three tasks. VRB2 and EgoHOS-Point each
produce two feasible trajectories and succeed on one task. The five-task checkpoint
succeeds on both executed trials, but four other trajectories are rejected. The 30-task
checkpoint therefore provides the strongest robot result despite the five-task
checkpoint having the best aggregate 2D external score.
Table~\ref{tab:robot_summary_v1} summarizes pre-repair invalid frames and
end-to-end task success.

\begin{table}[H]
\centering
\footnotesize
\setlength{\tabcolsep}{3pt}
\caption{Robot outcomes over six robot evaluation sequences per method and 210
input frames. Invalid is the pre-repair union of missing predictions, bad
depth, and geometric or workspace errors. Rejected trajectories count as failures.}
\label{tab:robot_summary_v1}
\begin{tabular}{@{}lrr@{}}
\toprule
Method & Invalid $\downarrow$ & Task success $\uparrow$ \\
\midrule
VRB2 & 72 (34.3\%) & 1/6 (16.7\%) \\
EgoHOS-Point & 70 (33.3\%) & 1/6 (16.7\%) \\
HOC2R fixed-two (5 tasks) & 42 (20.0\%) & 2/6 (33.3\%) \\
HOC2R fixed-two (30 tasks) & \textbf{30 (14.3\%)} &
\textbf{3/6 (50.0\%)} \\
\bottomrule
\end{tabular}
\end{table}

HOC2R succeeds in grasp-and-place, pushing, and two-hand pen-around-obstacle
tasks. One push reaches its endpoint but loses contact and is
counted as a failure. The difference between the five-task 2D result and the
30-task robot result shows that aggregate image localization alone does not
fully predict depth validity or trajectory feasibility. These executions establish
that contact trajectories can become executable post-contact motion
cues. Per-trajectory validity diagnostics and outcomes are reported in
Appendix~\ref{app:robot_outcomes_v1}, while all robot execution frames appear
in Appendix~\ref{app:robot_execution_frames_v1}.

\section{Discussion and Limitations}

\paragraph{Practical use.}
HOC2R is intended as a contact-trajectory extractor for downstream robot
learning. The released 30-task fixed-two checkpoint is the appropriate default
when a known active contact segment is available because it gives the highest
trajectory feasibility and task success. For a new environment, our data-efficiency
results motivate post-training on a small set selected for diversity across
hands, objects, viewpoints, and actions. This is a promising adaptation
strategy rather than evidence for a universal five-task recipe.

\paragraph{Where HOC2R helps.}
The clearest advantage is hand-object association in the evaluated multi-hand
scenes. HOC2R can preserve candidate regions around multiple hand-object
interactions while assigning stronger proposal scores to the contact associated
with the demonstrated manipulation. In contrast, the adapted
baselines can be distracted by resting hands, background hands, or unrelated
nearby objects. HOC2R also remains effective under the evaluated hand-color
shift. HOC2R transfers from blue hands in the annotated
benchmark to natural skin in the external RGB-D rollouts without changing its
inference procedure. In comparison, VRB2 requires a different skin-color
range, while EgoHOS-Point produces no hand predictions on the blue-hand
frames. In ordinary single-hand scenes, HOC2R is broadly competitive with
VRB2, but it does not outperform the baseline for every interaction pattern.

\paragraph{Limitations and scope.}
Our conclusions are subject to several limitations. The annotated benchmark
is a small dataset with 210 frames from 30 tasks, and the external
study uses one camera system and seven rollouts. The representation is limited
to two-finger interactions, and the model predicts each frame independently. RGB-D lifting
can fail through depth noise, calibration error, or a midpoint that samples
the wrong surface. Robot evaluation covers six sequences that begin
from established gripper contact and use open-loop inverse kinematics. VRB2 and
EgoHOS-Point are point-output adaptations rather than evaluations of their
original objectives. These constraints bound our claims to contact
localization and post-contact motion cues rather than autonomous grasping or
end-to-end imitation learning.

\section{Conclusion}

We presented HOC2R, a contact-trajectory extraction framework that connects
observed human demonstrations to post-contact robot motion. Frozen DINOv3 and
SAM~2.1 features support four lightweight heads and two deployment modes.
Count-aware automation gives the strongest contact-presence, contact-count,
and localization results on held-out in-domain tasks, while fixed-two localization transfers more
reliably when the active contact segment is known. The five-task fixed-two
checkpoint improves aggregate external localization over both adapted
baselines, while the 30-task checkpoint produces more feasible and successful
trajectories in the robot study.

The main finding is that HOC2R learns useful contact localization from
limited point annotations, transfers across the evaluated hand-color shift, and
shows its clearest advantage under multi-hand ambiguity. HOC2R provides a practical starting checkpoint for
extracting contact trajectories and pseudo-labeling human demonstrations.
Reliable large-scale use will require broader data, temporal contact
segmentation, uncertainty-aware filtering, and evaluation across more
environments and manipulation patterns.

\section*{Acknowledgments}
We thank Zekun Wang for support with data collection and annotation,
Linsen Wang for support during robot experiments, Sijie Xu for idea
discussions and annotation, and Olivia Xu and Runyu Tian for annotation.

\bibliography{references}

@article{xie2025human2robot,
  title   = {Human2Robot: Learning Robot Actions from Paired Human-Robot Videos},
  author  = {Xie, Sicheng and Cao, Haidong and Weng, Zejia and Xing, Zhen and Shen, Shiwei and Leng, Jiaqi and Qiu, Xipeng and Fu, Yanwei and Wu, Zuxuan and Jiang, Yu-Gang},
  journal = {arXiv preprint arXiv:2502.16587},
  year    = {2025}
}

@article{simeoni2025dinov3,
  title   = {{DINOv3}},
  author  = {Sim{\'e}oni, Oriane and Vo, Huy V. and Seitzer, Maximilian and Baldassarre, Federico and Oquab, Maxime and others},
  journal = {arXiv preprint arXiv:2508.10104},
  year    = {2025}
}

@article{ravi2024sam2,
  title   = {{SAM 2}: Segment Anything in Images and Videos},
  author  = {Ravi, Nikhila and Gabeur, Valentin and Hu, Yuan-Ting and Hu, Ronghang and Ryali, Chaitanya and others},
  journal = {arXiv preprint arXiv:2408.00714},
  year    = {2024}
}

@inproceedings{bahl2023affordances,
  title     = {Affordances from Human Videos as a Versatile Representation for Robotics},
  author    = {Bahl, Shikhar and Mendonca, Russell and Chen, Lili and Jain, Unnat and Pathak, Deepak},
  booktitle = {Proceedings of the IEEE/CVF Conference on Computer Vision and Pattern Recognition},
  year      = {2023}
}

@inproceedings{shan2020hands,
  title     = {Understanding Human Hands in Contact at Internet Scale},
  author    = {Shan, Dandan and Geng, Jiaqi and Shu, Michelle and Fouhey, David F.},
  booktitle = {Proceedings of the IEEE/CVF Conference on Computer Vision and Pattern Recognition},
  year      = {2020}
}

@inproceedings{zhang2022egohos,
  title     = {Fine-Grained Egocentric Hand-Object Segmentation: Dataset, Model, and Applications},
  author    = {Zhang, Lingzhi and Zhou, Shenghao and Stent, Simon and Shi, Jianbo},
  booktitle = {Proceedings of the European Conference on Computer Vision},
  year      = {2022}
}

@inproceedings{mandlekar2021what,
  title     = {What Matters in Learning from Offline Human Demonstrations for Robot Manipulation},
  author    = {Mandlekar, Ajay and Xu, Danfei and Wong, Josiah and Nasiriany, Soroush and Wang, Chen and Kulkarni, Rohun and Fei-Fei, Li and Savarese, Silvio and Zhu, Yuke and Mart{\'i}n-Mart{\'i}n, Roberto},
  booktitle = {Proceedings of the Conference on Robot Learning},
  year      = {2021}
}

@inproceedings{wang2023mimicplay,
  title     = {{MimicPlay}: Long-Horizon Imitation Learning by Watching Human Play},
  author    = {Wang, Chen and Fan, Linxi and Sun, Jiankai and Zhang, Ruohan and Fei-Fei, Li and Xu, Danfei and Zhu, Yuke and Anandkumar, Anima},
  booktitle = {Proceedings of the Conference on Robot Learning},
  year      = {2023}
}

@inproceedings{yan2023imitationnet,
  title     = {{ImitationNet}: Unsupervised Human-to-Robot Motion Retargeting via Shared Latent Space},
  author    = {Yan, Yashuai and Valls Mascaro, Esteve and Lee, Dongheui},
  booktitle = {Proceedings of the IEEE-RAS International Conference on Humanoid Robots},
  year      = {2023}
}

@inproceedings{do2018affordancenet,
  title     = {{AffordanceNet}: An End-to-End Deep Learning Approach for Object Affordance Detection},
  author    = {Do, Thanh-Toan and Nguyen, Anh and Reid, Ian},
  booktitle = {Proceedings of the IEEE International Conference on Robotics and Automation},
  year      = {2018}
}

@article{shi2026howtransfer,
  title   = {Hand-Centric Human-to-Robot Trajectory Transfer from Video Demonstrations via Open-World Contact Localization},
  author  = {Shi, Yitian and Wen, Di and Han, Zhengqi and Guo, Zicheng and Hu, Yu and Welte, Edgar and Peng, Kunyu and Stiefelhagen, Rainer and Rayyes, Rania},
  journal = {arXiv preprint arXiv:2606.10743},
  year    = {2026}
}

@article{azirar2026contactflow,
  title   = {{ContactFlow}: A Video Action Conditioning That Transfers across Embodiments},
  author  = {Azirar, Sami and Pallotta, Enrico and Nogga, Jan and Gall, J{\"u}rgen and Behnke, Sven and Blum, Hermann},
  journal = {arXiv preprint arXiv:2607.26579},
  year    = {2026}
}

@inproceedings{wen2024atm,
  title     = {Any-Point Trajectory Modeling for Policy Learning},
  author    = {Wen, Chuan and Lin, Xingyu and So, John and Chen, Kai and Dou, Qi and Gao, Yang and Abbeel, Pieter},
  booktitle = {Proceedings of Robotics: Science and Systems},
  year      = {2024}
}

@inproceedings{nair2023r3m,
  title     = {{R3M}: A Universal Visual Representation for Robot Manipulation},
  author    = {Nair, Suraj and Rajeswaran, Aravind and Kumar, Vikash and Finn, Chelsea and Gupta, Abhinav},
  booktitle = {Proceedings of the Conference on Robot Learning},
  year      = {2023}
}

@inproceedings{majumdar2023vc1,
  title     = {Where Are We in the Search for an Artificial Visual Cortex for Embodied Intelligence?},
  author    = {Majumdar, Arjun and Yadav, Karmesh and Arnaud, Sergio and Ma, Yecheng Jason and Chen, Claire and Silwal, Sneha and Jain, Aryan and Berges, Vincent-Pierre and Wu, Tingfan and Vakil, Jay and Abbeel, Pieter and Malik, Jitendra and Batra, Dhruv and Lin, Yixin and Maksymets, Oleksandr and Rajeswaran, Aravind and Meier, Franziska},
  booktitle = {Advances in Neural Information Processing Systems},
  year      = {2023}
}
\bibliographystyle{named}

\appendix

\section{Contact-Count Diagnostics}
\label{app:contact_count_acra}

\begin{figure}[H]
\centering
\begin{minipage}{0.49\linewidth}\centering
\includegraphics[width=\linewidth]{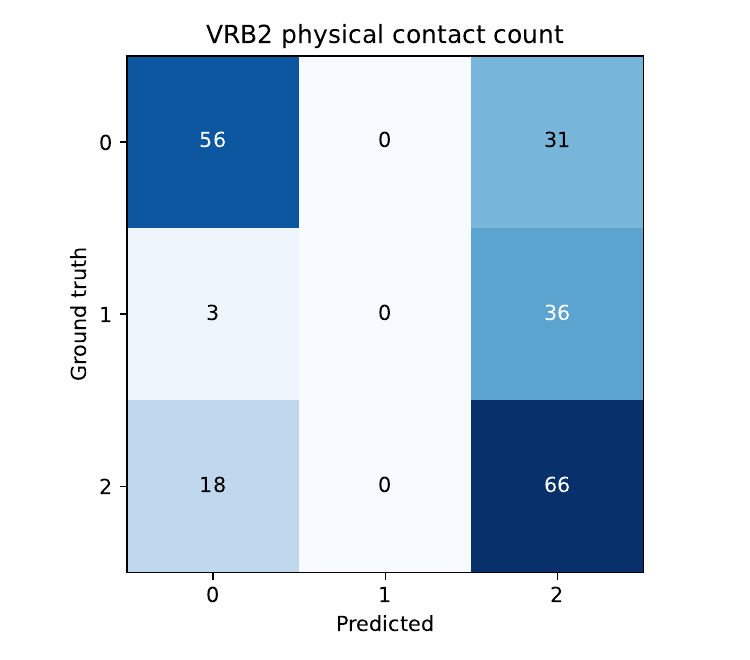}\\[-2pt]
\small VRB2
\end{minipage}\hfill
\begin{minipage}{0.49\linewidth}\centering
\includegraphics[width=\linewidth]{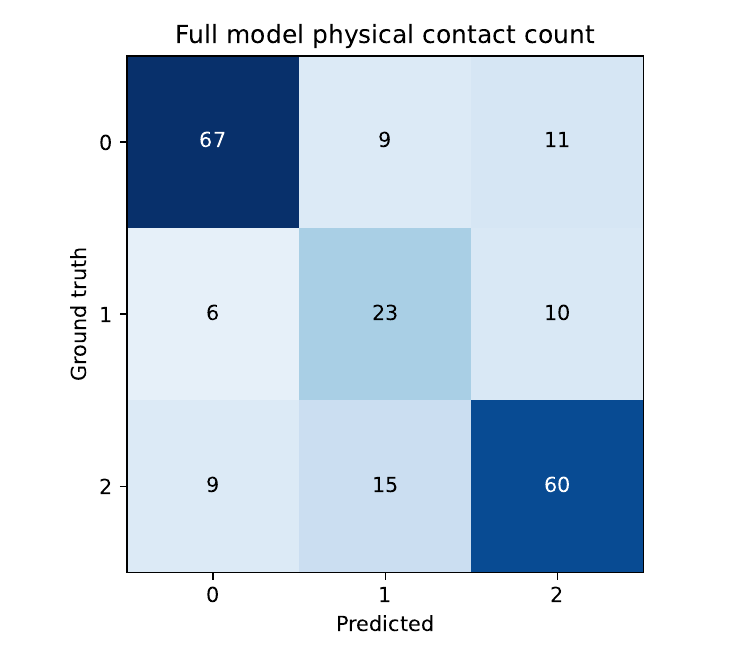}\\[-2pt]
\small HOC2R count-aware
\end{minipage}
\caption{In-domain contact-count confusion over 210 held-out frames. Rows show
ground truth, and columns show predictions.}
\label{fig:count_confusion_v1}
\end{figure}

\paragraph{Contact-count confusion.}
Neither adapted baseline explicitly estimates the contact count. VRB2 emits two
Gaussian mixture centers when valid intersection evidence exists, while EgoHOS-Point emits
two component centers when a contact boundary is available. Otherwise, they
return no points. HOC2R count-aware automation is the only compared method with
an explicit contact-presence and contact-count head. Figure~\ref{fig:count_confusion_v1}
shows its resulting predictions alongside the VRB2 adapter. Count-aware
automation reaches 87.8\% contact-presence recall, detecting contact in 108 of 123 positive
frames. Its mean-class accuracy, computed as the unweighted mean of the three
contact-count recalls, is 69.1\%. Recall for the no-contact,
one-visible-point, and two-visible-point classes is 77.0\%, 59.0\%, and 71.4\%,
respectively.
EgoHOS-Point has no in-domain confusion matrix because its pretrained hand
detector returns no prediction for any blue-hand frame.

\section{External Per-Rollout Localization}
\label{app:external_per_rollout_acra}

This section reports the conditional mean localization distance for each of the
seven external RGB-D rollouts.

\begin{table}[H]
\centering
\scriptsize
\setlength{\tabcolsep}{3pt}
\caption{External per-rollout conditional mean localization distance in pixels.
Lower is better. EgoHOS denotes EgoHOS-Point; HOC2R-30 and HOC2R-5 denote
fixed-two checkpoints trained on 30 and 5 tasks, respectively.}
\label{tab:external_tasks_acra}
\begin{tabular}{@{}lrrrr@{}}
\toprule
Task & VRB2 $\downarrow$ & EgoHOS $\downarrow$ &
HOC2R-30 $\downarrow$ & HOC2R-5 $\downarrow$ \\
\midrule
Biscuit & 20.84 & 21.36 & 24.88 & \textbf{18.75} \\
Cube & 22.63 & 46.47 & 15.79 & \textbf{6.19} \\
Fruit knife & \textbf{26.13} & 133.78 & 65.24 & 38.44 \\
Pencil (2 hands) & 165.58 & 124.09 & 30.48 & \textbf{24.18} \\
Push case & \textbf{21.52} & 33.14 & 68.65 & 70.27 \\
Spoon & 28.27 & 80.99 & 35.14 & \textbf{21.48} \\
Tissue & \textbf{25.14} & 62.87 & 43.66 & 25.92 \\
\bottomrule
\end{tabular}
\end{table}

\section{RGB-D Calibration and Trajectory Safety}
\label{app:rgbd_robot_acra}

\paragraph{Calibration.}
For image midpoint $p_m=(u_m,v_m)$, let $z$ be the median valid aligned depth
in its $3\times3$ neighborhood and $K$ the RealSense color-camera intrinsic
matrix. The camera-frame and robot-base points are
\begin{equation}
    p_C=zK^{-1}[u_m,v_m,1]^\top,
    \qquad
    \widetilde p_B=T_{B\leftarrow C}\widetilde p_C,
\end{equation}
where tildes denote homogeneous coordinates. We estimate the fixed
camera-to-robot transform with the Daniilidis hand-eye solver using ten
ChArUco poses and check it on five additional poses. Every method uses the
same calibration.

\paragraph{Invalid-point detection.}
Using the fixed tool orientation and table configuration, we manually
validated the Kinova workspace as $x\in[0.25,0.73]$ m,
$y\in[-0.33,0.33]$ m, and $z\in[0.016,0.40]$ m. A frame is invalid when it
has no 2D prediction, no usable aligned depth, or a point outside this
workspace. Depth values below $-0.03$ m are catastrophic errors. We also flag
a depth that differs from the mean of two mutually consistent neighbors by
more than $0.10$ m, or a 3D point whose residual from its interpolated neighbor
position exceeds $0.06$ m when the surrounding displacement is at most
$0.10$ m.

\begin{table}[H]
\centering
\footnotesize
\setlength{\tabcolsep}{3pt}
\caption{Shared trajectory-repair limits. Run lengths are measured in frames.
Exceeding any limit rejects the trajectory.}
\label{tab:trajectory_filter_limits_acra}
\begin{tabular}{@{}lrrr@{}}
\toprule
Error & Fraction & Internal & Endpoint \\
\midrule
Missing 2D & 25\% & 5 & 3 \\
Depth/geometry/workspace & 15\% & 3 & 1 \\
\bottomrule
\end{tabular}
\end{table}

\paragraph{Repair and rejection.}
Table~\ref{tab:trajectory_filter_limits_acra} gives the maximum repairable
fractions and consecutive runs. Short internal gaps are linearly interpolated,
and admissible endpoint gaps use the nearest valid point. A combined invalid
run longer than five frames also causes rejection. After repair, trajectories with a
remaining step above $0.10$ m are rejected, while accepted trajectories are resampled
to at most $0.02$ m waypoint spacing.

\section{Robot Evaluation Details}

\subsection{Trajectory Validity and Robot Outcomes}
\label{app:robot_outcomes_v1}

We denote the paired evaluation sequences as Grasp 1/2, Push 1/2, and Pen 1/2.
Table~\ref{tab:robot_filter_causes_v1} reports pre-repair errors. Missing 2D
measures detector dropout, bad depth denotes unusable aligned depth, and
geometry/workspace errors include
discontinuities and robot-base points outside the validated workspace.
``Invalid'' is the union of these categories, so overlapping causes are counted
once. The fixed-two decoder emits two candidates for every frame
in a known active contact segment. In contrast, VRB2 and EgoHOS-Point can return no
point when their upstream hand-object intersection or contact-boundary
prediction is empty. Missing 2D therefore measures upstream coverage, not
contact-presence accuracy. HOC2R trajectories can still be
invalid because of depth, geometry, or workspace errors, and are rejected when
these errors exceed the repair limits in
Appendix~\ref{app:rgbd_robot_acra}. Table~\ref{tab:robot_invalid_per_demo_v1}
breaks the invalid frames down by evaluation sequence.

\begin{table}[H]
\centering
\small
\caption{Pre-repair diagnostics across the six robot evaluation sequences and 210
input frames per method. Lower is better.}
\label{tab:robot_filter_causes_v1}
\resizebox{\columnwidth}{!}{%
\begin{tabular}{@{}lrrrr@{}}
\toprule
Method & Missing 2D $\downarrow$ & Bad depth $\downarrow$ &
Geometry/workspace $\downarrow$ & Invalid union $\downarrow$ \\
\midrule
VRB2 & 54 & \textbf{0} & \textbf{18} & 72 \\
EgoHOS-Point & 28 & 1 & 47 & 70 \\
HOC2R fixed-two (5 tasks) & \textbf{0} & 3 & 41 & 42 \\
HOC2R fixed-two (30 tasks) & \textbf{0} & 5 & 27 & \textbf{30} \\
\bottomrule
\end{tabular}%
}
\end{table}

\begin{table}[H]
\centering
\small
\caption{Invalid pre-repair frames per robot evaluation sequence. Lower is better.}
\label{tab:robot_invalid_per_demo_v1}
\resizebox{\columnwidth}{!}{%
\begin{tabular}{@{}lrrrrrr@{}}
\toprule
& \multicolumn{6}{c}{Invalid frames $\downarrow$} \\
\cmidrule(l){2-7}
Method & \shortstack{Grasp 1\\(30)} & \shortstack{Grasp 2\\(34)} &
\shortstack{Push 1\\(30)} & \shortstack{Push 2\\(25)} &
\shortstack{Pen 1\\(47)} & \shortstack{Pen 2\\(44)} \\
\midrule
VRB2 & 19 & 8 & 9 & 6 & 17 & 13 \\
EgoHOS-Point & \textbf{0} & 4 & 24 & 15 & 11 & 16 \\
HOC2R fixed-two (5 tasks) & 8 & \textbf{0} & 10 & \textbf{1} & 11 & 12 \\
HOC2R fixed-two (30 tasks) & 9 & 1 & \textbf{3} & \textbf{1} & \textbf{6} & \textbf{10} \\
\bottomrule
\end{tabular}%
}
\end{table}

\begin{table}[H]
\centering
\small
\caption{Individual robot outcomes. \textbf{S} denotes an executed task
success, F an executed task failure, and R a trajectory rejected before execution by
the shared feasibility filter.}
\label{tab:robot_outcomes_v1}
\resizebox{\columnwidth}{!}{%
\begin{tabular}{@{}lcccccc@{}}
\toprule
Method & Grasp 1 & Grasp 2 & Push 1 & Push 2 & Pen 1 & Pen 2 \\
\midrule
VRB2 & R & \textbf{S} & R & F & R & R \\
EgoHOS-Point & F & \textbf{S} & R & R & R & R \\
HOC2R fixed-two (5 tasks) & R & \textbf{S} & R & \textbf{S} & R & R \\
HOC2R fixed-two (30 tasks) & R & \textbf{S} & F & \textbf{S} & \textbf{S} & R \\
\bottomrule
\end{tabular}%
}
\end{table}

Among the executed failures, the VRB2 Push 2 trajectory makes a large lateral
deviation mid-motion, after which the case does not reach the target.
EgoHOS-Point executes Grasp 1 but finishes approximately 4 cm outside the
basket. The 30-task HOC2R Push 1 trajectory reaches the endpoint but briefly
loses contact, so the task is marked as failed. All other planned trajectories
were withheld by the shared pre-execution safety criteria. This
distinguishes an executed task failure from a trajectory rejected before robot
motion.

\subsection{Robot Execution Frames}
\label{app:robot_execution_frames_v1}

\begin{figure}[H]
\centering
\setlength{\tabcolsep}{1pt}
\begin{tabular}{@{}cccc@{}}
\multicolumn{4}{@{}p{\columnwidth}@{}}{\textbf{VRB2, Grasp 2, reached basket, task success}} \\[-2pt]
\includegraphics[width=.24\linewidth]{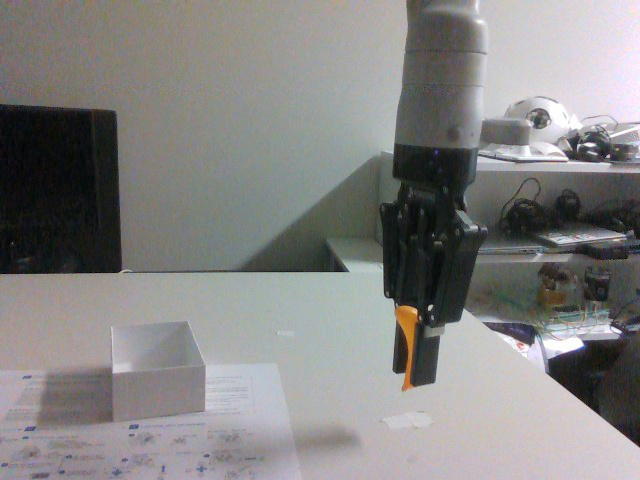} &
\includegraphics[width=.24\linewidth]{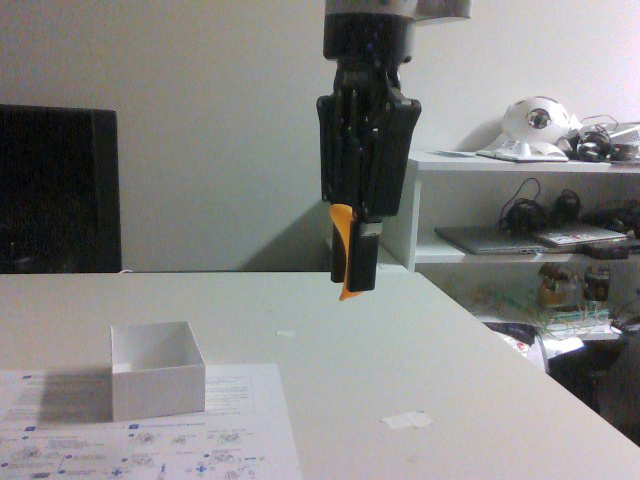} &
\includegraphics[width=.24\linewidth]{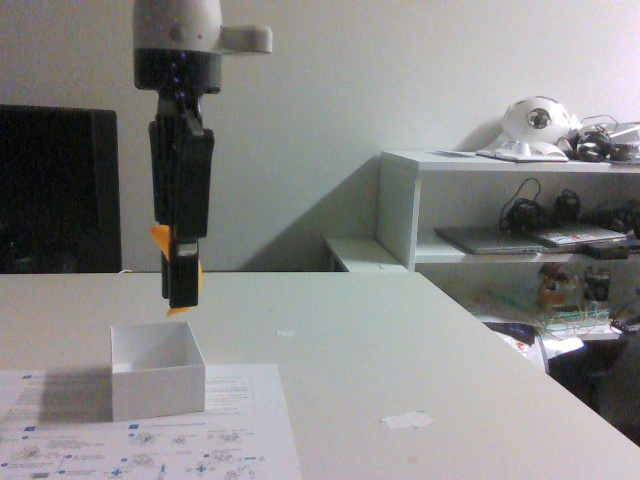} &
\includegraphics[width=.24\linewidth]{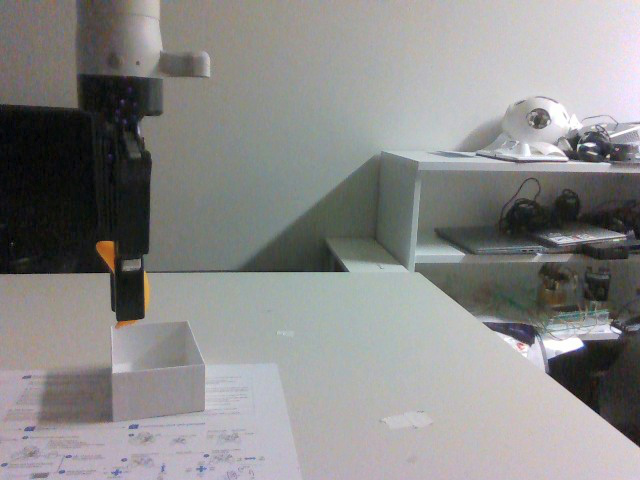} \\[5pt]
\multicolumn{4}{@{}p{\columnwidth}@{}}{\textbf{VRB2, Push 2, large lateral deviation, task failure}} \\[-2pt]
\includegraphics[width=.24\linewidth]{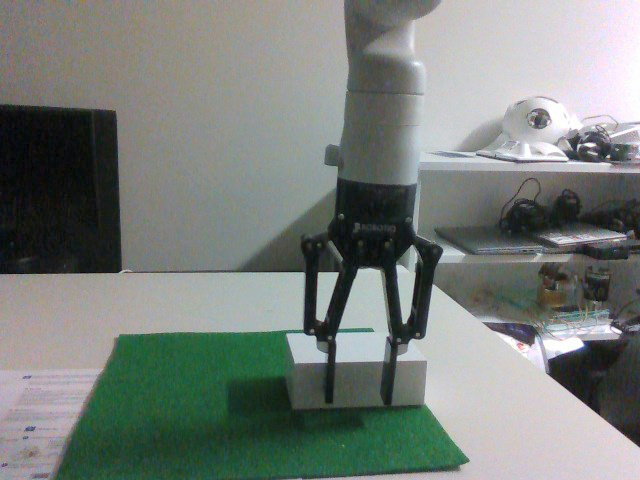} &
\includegraphics[width=.24\linewidth]{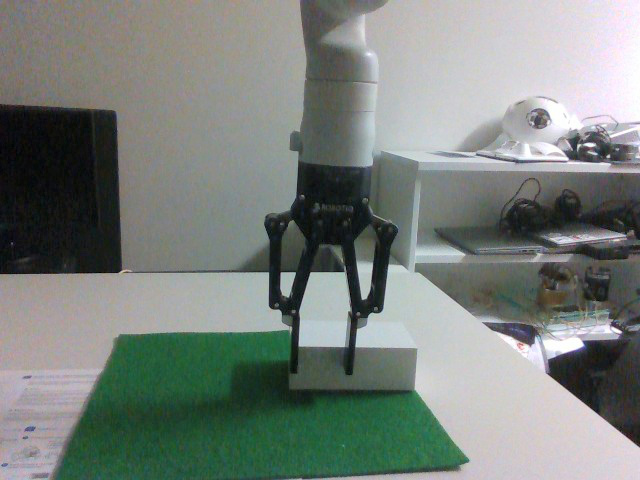} &
\includegraphics[width=.24\linewidth]{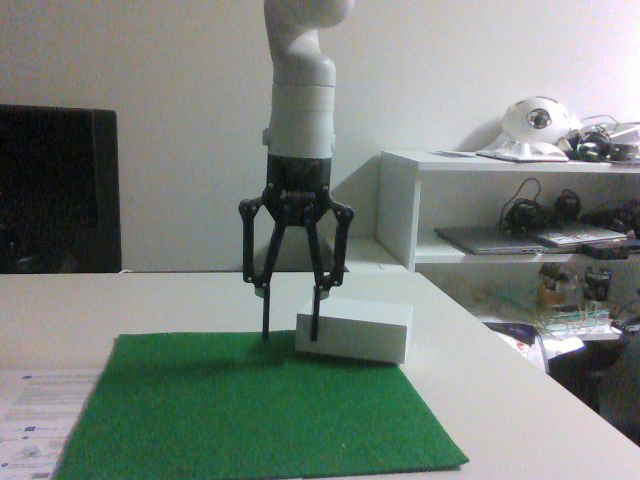} &
\includegraphics[width=.24\linewidth]{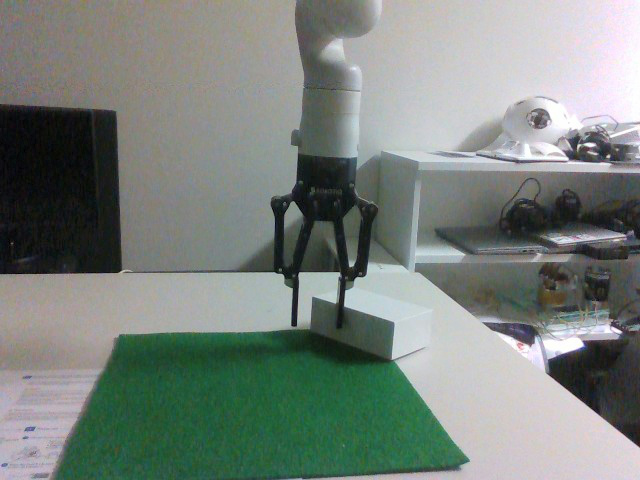}
\end{tabular}
\caption{Robot executions for VRB2. Four selected frames are shown for
each executed trajectory. Grasp 2 is a task success, while Push 2 is a task
failure because a large lateral deviation prevents the case from reaching the
target.}
\label{fig:robot_frames_vrb_v1}
\end{figure}

\begin{figure}[H]
\centering
\setlength{\tabcolsep}{1pt}
\begin{tabular}{@{}cccc@{}}
\multicolumn{4}{@{}p{\columnwidth}@{}}{\textbf{EgoHOS-Point, Grasp 1, ended 4 cm outside basket, task failure}} \\[-2pt]
\includegraphics[width=.24\linewidth]{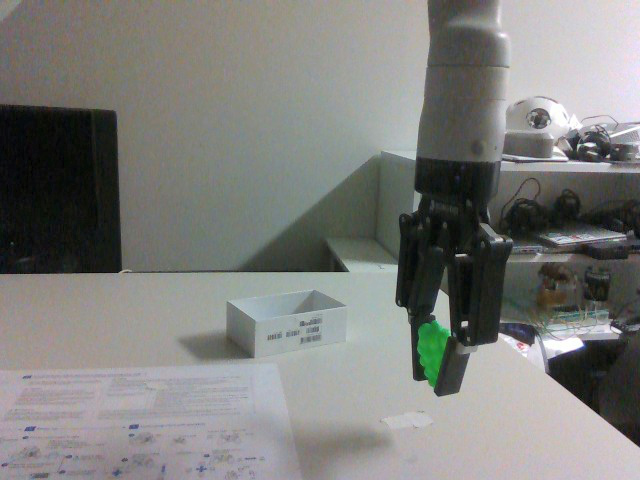} &
\includegraphics[width=.24\linewidth]{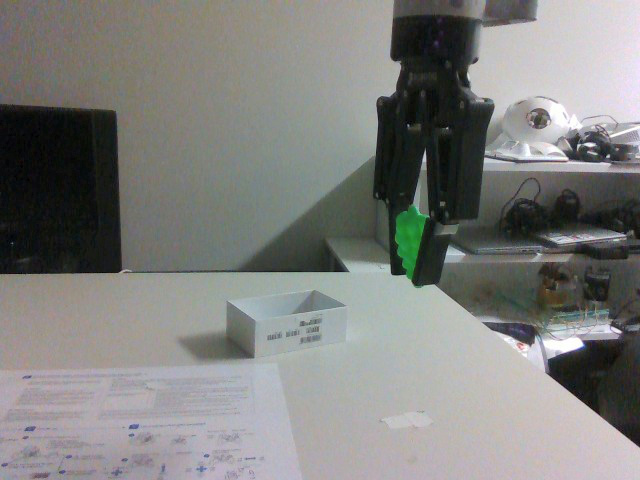} &
\includegraphics[width=.24\linewidth]{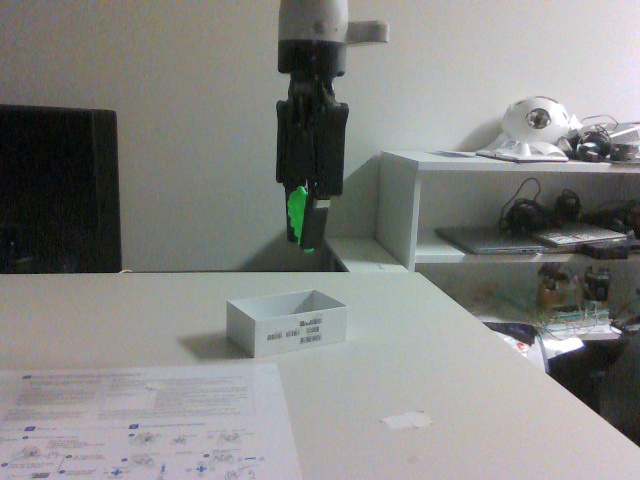} &
\includegraphics[width=.24\linewidth]{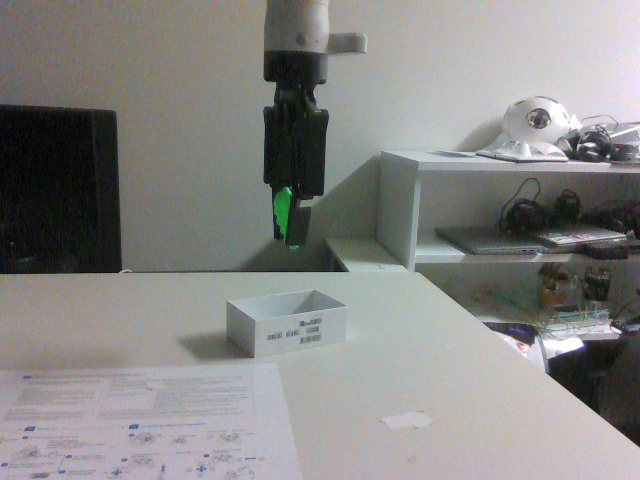} \\[5pt]
\multicolumn{4}{@{}p{\columnwidth}@{}}{\textbf{EgoHOS-Point, Grasp 2, reached basket, task success}} \\[-2pt]
\includegraphics[width=.24\linewidth]{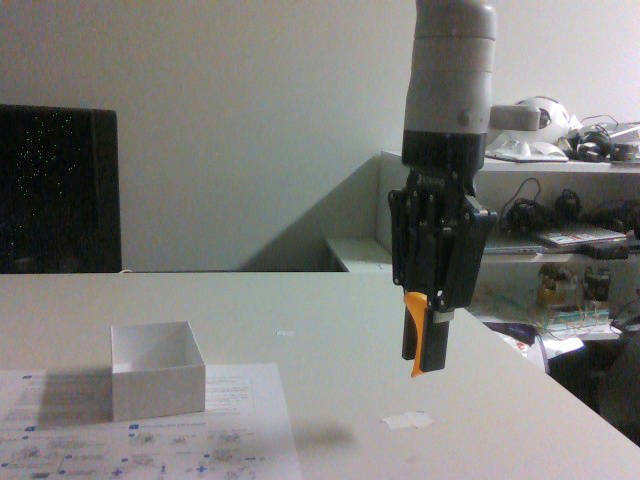} &
\includegraphics[width=.24\linewidth]{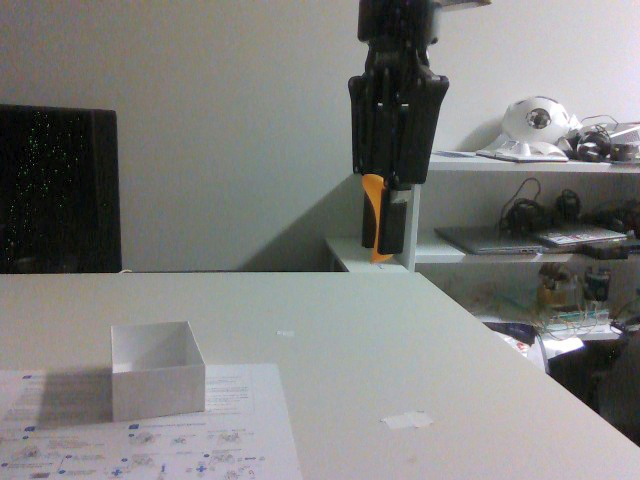} &
\includegraphics[width=.24\linewidth]{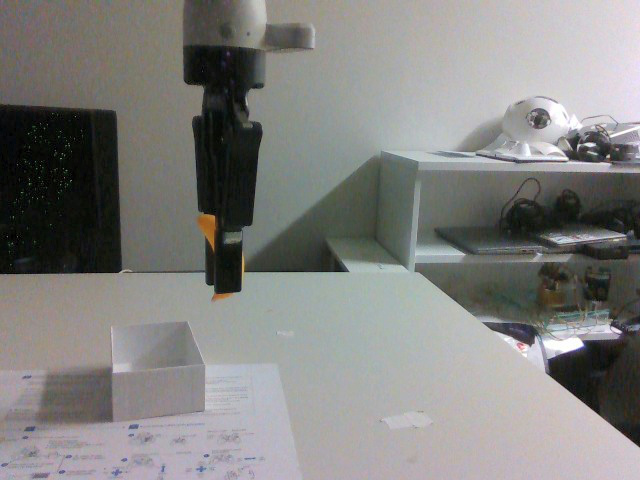} &
\includegraphics[width=.24\linewidth]{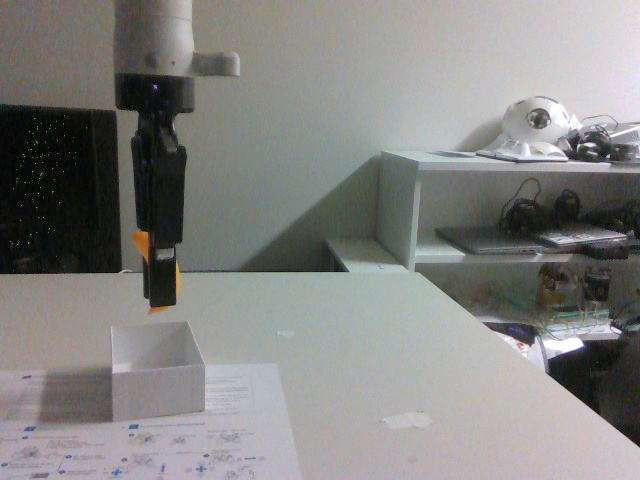}
\end{tabular}
\caption{Robot executions for EgoHOS-Point. Four selected frames are shown
for each executed trajectory. Grasp 1 is a task failure after ending outside
the basket, while Grasp 2 is a task success.}
\label{fig:robot_frames_egohos_v1}
\end{figure}

\begin{figure}[H]
\centering
\setlength{\tabcolsep}{1pt}
\begin{tabular}{@{}cccc@{}}
\multicolumn{4}{@{}p{\columnwidth}@{}}{\textbf{HOC2R fixed-two (5 tasks), Grasp 2, reached basket, task success}} \\[-2pt]
\includegraphics[width=.24\linewidth]{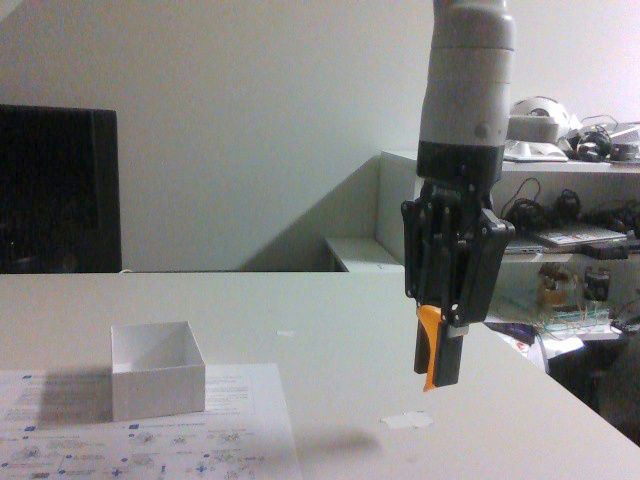} &
\includegraphics[width=.24\linewidth]{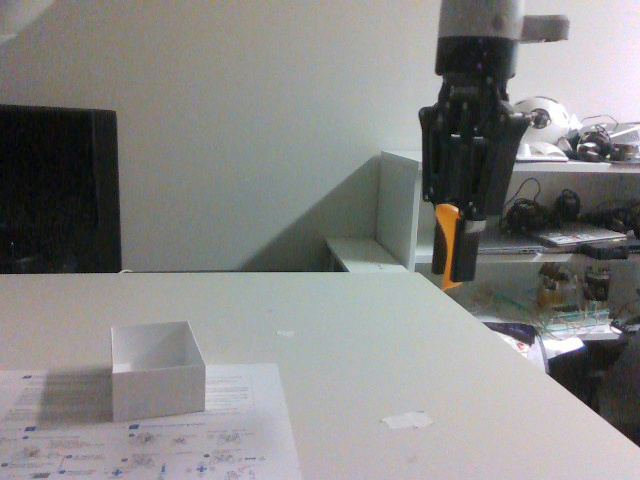} &
\includegraphics[width=.24\linewidth]{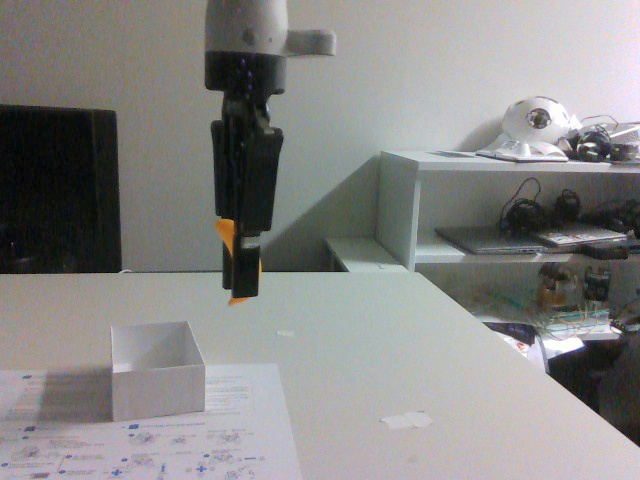} &
\includegraphics[width=.24\linewidth]{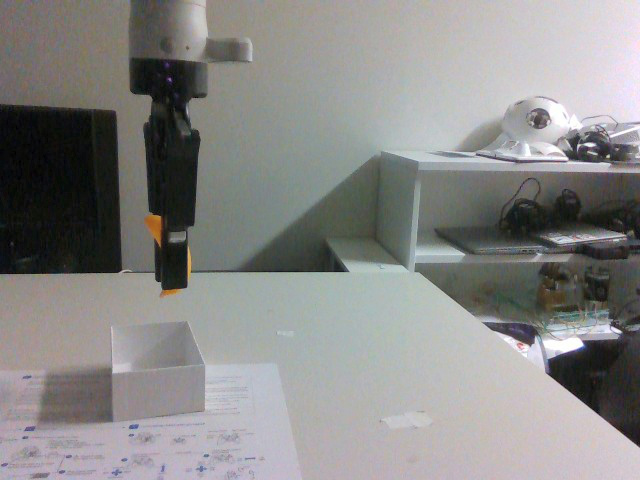} \\[5pt]
\multicolumn{4}{@{}p{\columnwidth}@{}}{\textbf{HOC2R fixed-two (5 tasks), Push 2, reached target, task success}} \\[-2pt]
\includegraphics[width=.24\linewidth]{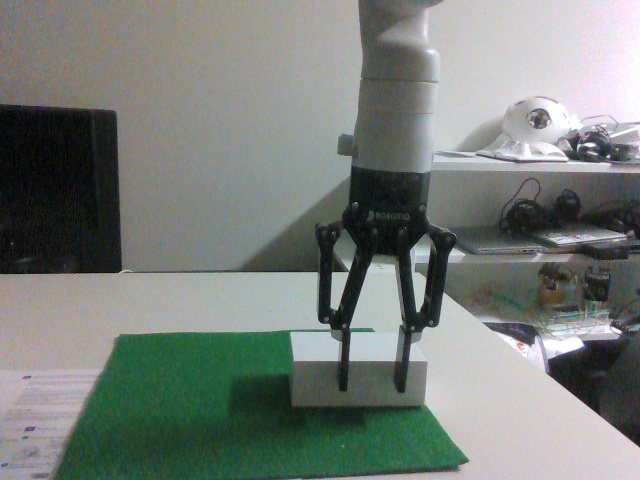} &
\includegraphics[width=.24\linewidth]{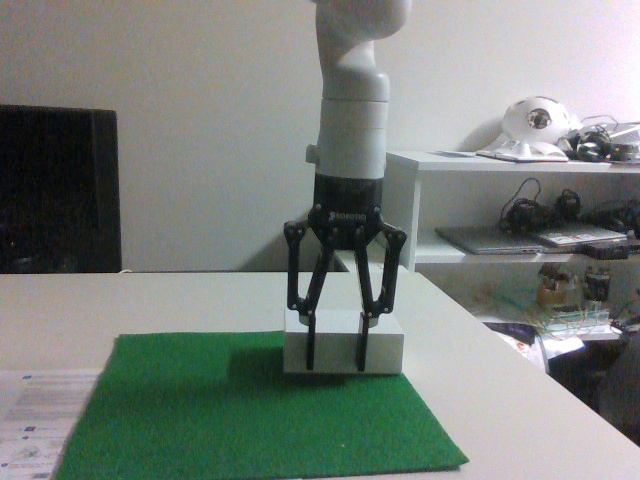} &
\includegraphics[width=.24\linewidth]{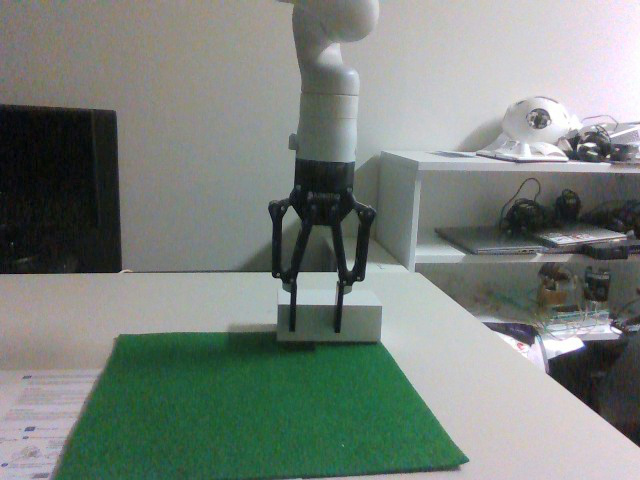} &
\includegraphics[width=.24\linewidth]{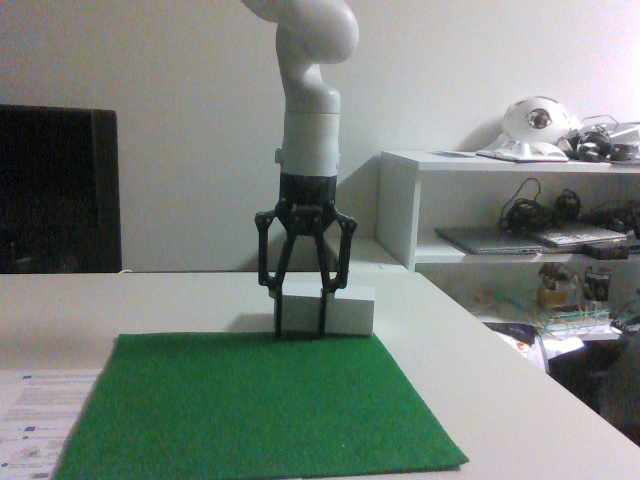}
\end{tabular}
\caption{Robot executions for HOC2R fixed-two (5 tasks). Four selected
frames are shown for each executed trajectory. Both feasible trajectories
succeed.}
\label{fig:robot_frames_hoc2r_5_v1}
\end{figure}

\begin{figure}[H]
\centering
\setlength{\tabcolsep}{1pt}
\begin{tabular}{@{}cccc@{}}
\multicolumn{4}{@{}p{\columnwidth}@{}}{\textbf{HOC2R fixed-two (30 tasks), Grasp 2, reached basket, task success}} \\[-2pt]
\includegraphics[width=.24\linewidth]{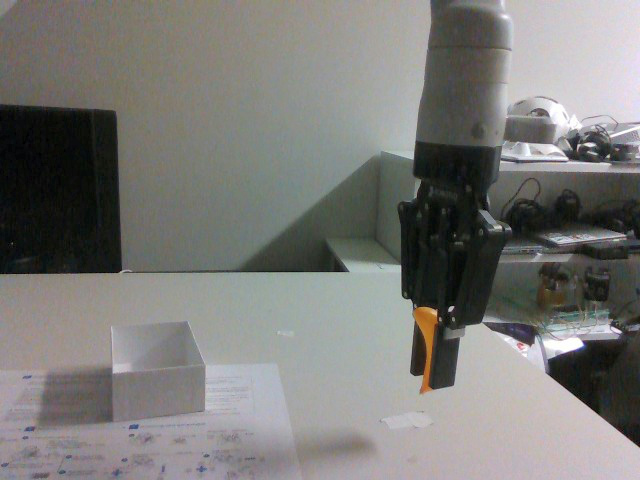} &
\includegraphics[width=.24\linewidth]{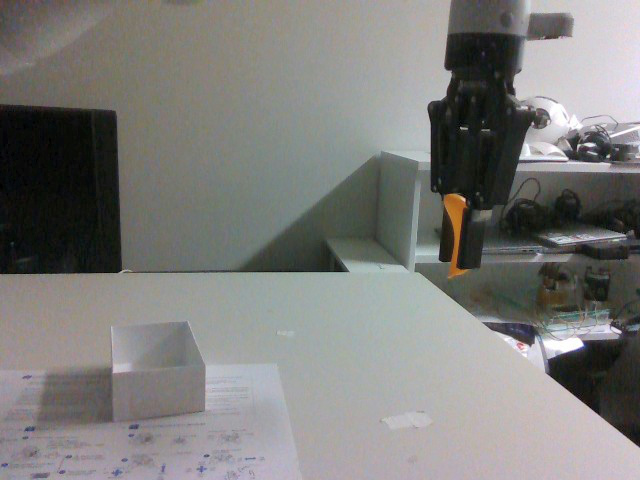} &
\includegraphics[width=.24\linewidth]{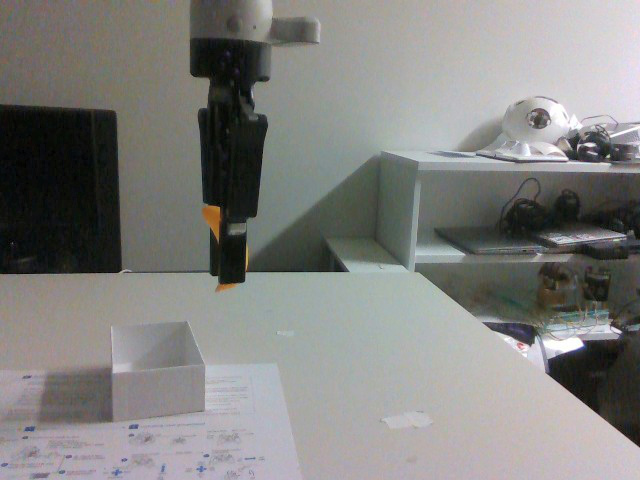} &
\includegraphics[width=.24\linewidth]{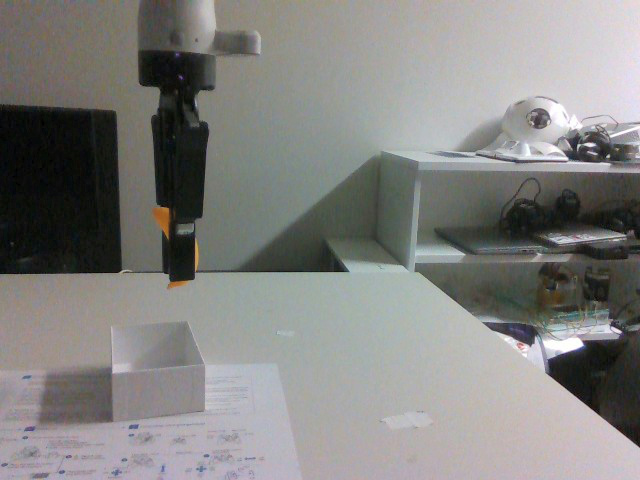} \\[5pt]
\multicolumn{4}{@{}p{\columnwidth}@{}}{\textbf{HOC2R fixed-two (30 tasks), Push 1, lost contact, task failure}} \\[-2pt]
\includegraphics[width=.24\linewidth]{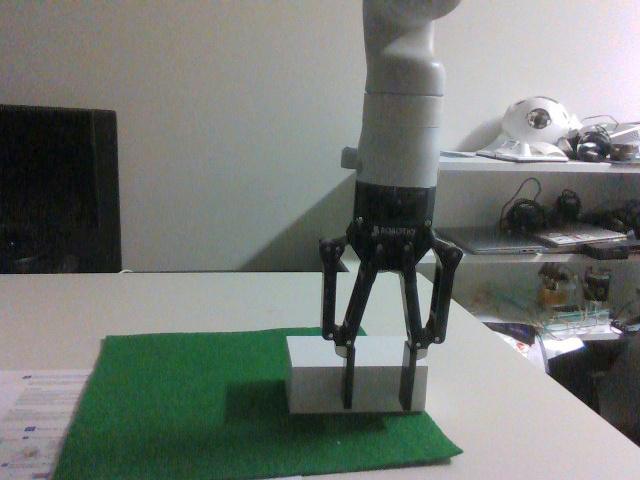} &
\includegraphics[width=.24\linewidth]{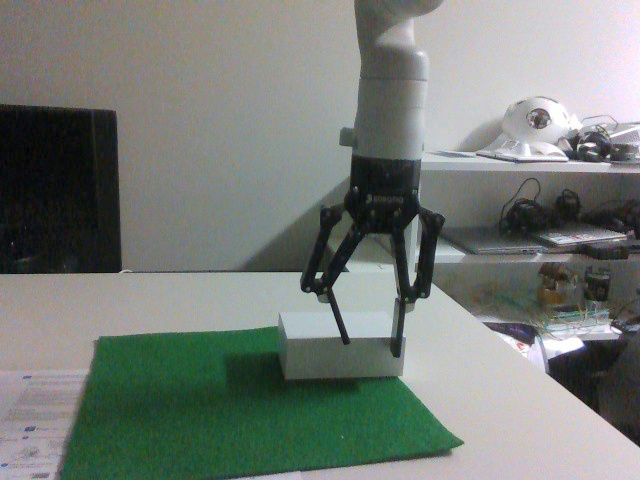} &
\includegraphics[width=.24\linewidth]{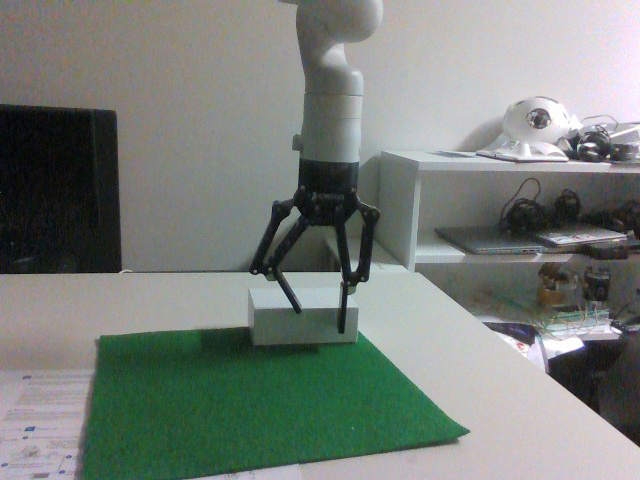} &
\includegraphics[width=.24\linewidth]{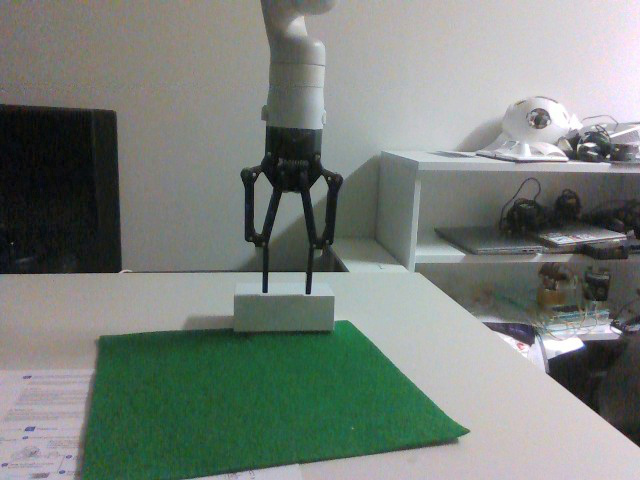} \\[5pt]
\multicolumn{4}{@{}p{\columnwidth}@{}}{\textbf{HOC2R fixed-two (30 tasks), Push 2, reached target, task success}} \\[-2pt]
\includegraphics[width=.24\linewidth]{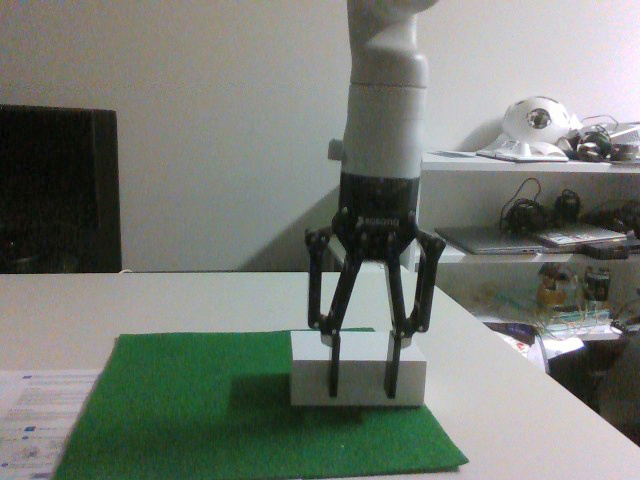} &
\includegraphics[width=.24\linewidth]{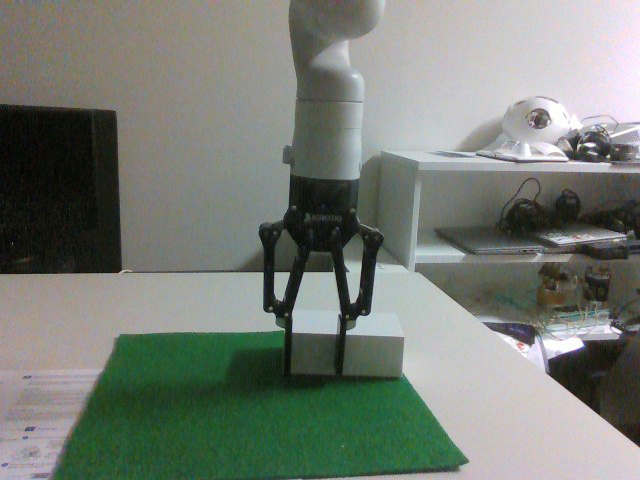} &
\includegraphics[width=.24\linewidth]{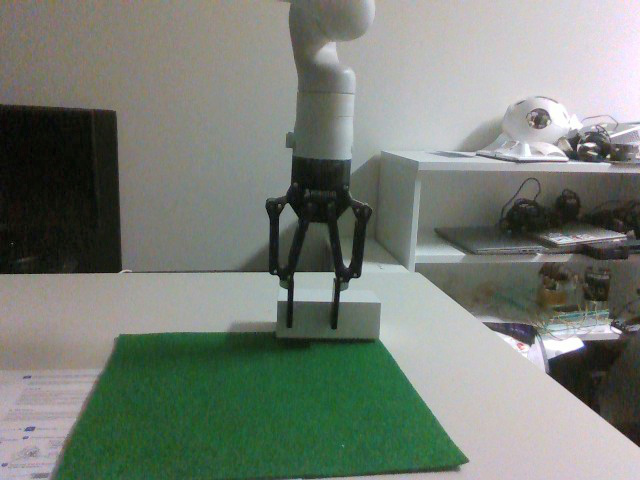} &
\includegraphics[width=.24\linewidth]{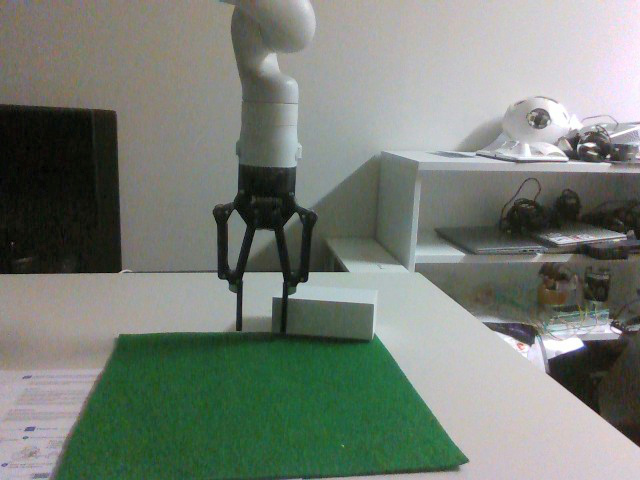} \\[5pt]
\multicolumn{4}{@{}p{\columnwidth}@{}}{\textbf{HOC2R fixed-two (30 tasks), Pen 1, moved pen around obstacle, task success}} \\[-2pt]
\includegraphics[width=.24\linewidth]{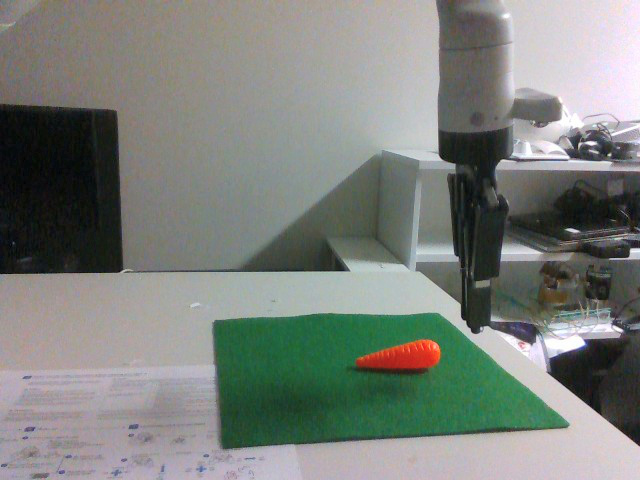} &
\includegraphics[width=.24\linewidth]{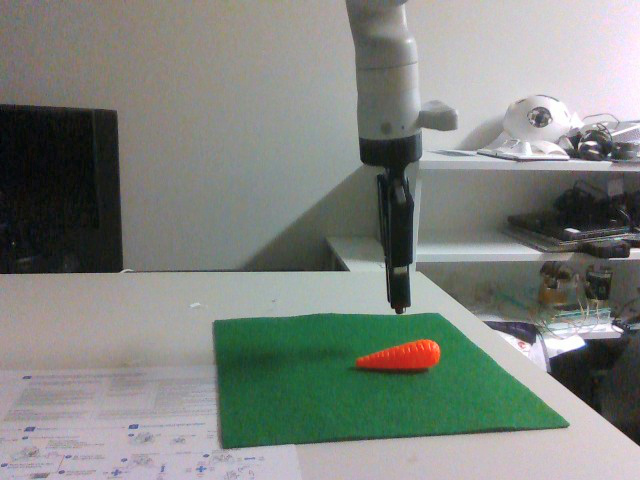} &
\includegraphics[width=.24\linewidth]{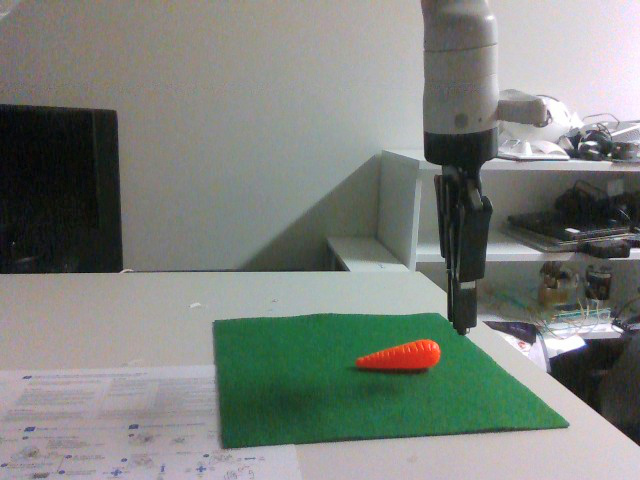} &
\includegraphics[width=.24\linewidth]{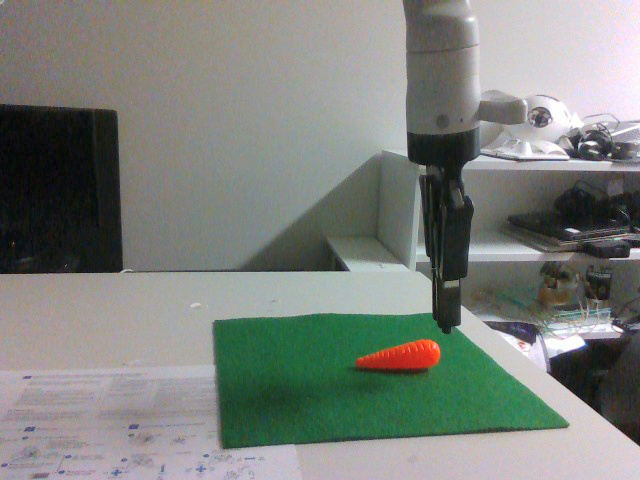}
\end{tabular}
\caption{Robot executions for HOC2R fixed-two (30 tasks). Four selected
frames are shown for each executed trajectory. Three of four trajectories
succeed. Push 1 reaches its endpoint but loses contact during motion.}
\label{fig:robot_frames_hoc2r_30_v1}
\end{figure}

Figures~\ref{fig:robot_frames_vrb_v1}--\ref{fig:robot_frames_hoc2r_30_v1} show
all ten trajectories that passed the shared feasibility filter and were
executed on the robot. Each row is labeled with the method, task, observed outcome, and
success state. Rejected trajectories are reported in
Table~\ref{tab:robot_outcomes_v1} but are not executed for safety.

\end{document}